\PassOptionsToPackage{table}{xcolor}
\documentclass[10pt,twocolumn,letterpaper]{article}

\usepackage[pagenumbers]{cvpr}

\usepackage{caption}
\usepackage{booktabs}
\usepackage{multirow}
\usepackage{adjustbox}
\usepackage{amsmath}
\usepackage{amssymb}
\usepackage{array}
\usepackage{wrapfig}
\usepackage{threeparttable}
\usepackage{enumitem}
\usepackage{placeins}
\usepackage[percent]{overpic}
\usepackage{titletoc}

\definecolor{SRTbg}{HTML}{FBFFF9}
\definecolor{SRTtext}{HTML}{82B366}

\newcommand{\SRTbadge}{%
  \colorbox{SRTbg}{\textcolor{SRTtext}{\textbf{SRT}\vphantom{A}}}%
}

\usepackage{tcolorbox}
\tcbuselibrary{skins,breakable}
\newtcolorbox{discussionbox}{
  breakable,
  colback=gray!8,
  colframe=gray!40,
  boxrule=0.5pt,
  arc=5pt,
  left=6pt,
  right=6pt,
  top=6pt,
  bottom=6pt
}

\usepackage{algorithm}
\usepackage{algpseudocode}
\usepackage{etoolbox}
\usepackage{xspace}

\renewcommand{\ie}{\textit{i.e.}\xspace}
\renewcommand{\eg}{\textit{e.g.}\xspace}

\renewcommand{\cf}{\textit{cf.}\xspace}

\makeatletter
\def\cvprsect#1{\cvprsection[#1]{\texorpdfstring{\hskip -1em.~}{}#1}}
\def\cvprsubsect#1{\cvprsubsection[#1]{\texorpdfstring{\hskip -1em.~}{}#1}}
\def\cvprsubsubsect#1{\cvprsubsubsection[#1]{\texorpdfstring{\hskip -1em.~}{}#1}}
\makeatother

\usepackage{utfsym}
\newcommand{\heart}{$\;\!$\usym{2665}}

\definecolor{cvprblue}{rgb}{0.21,0.49,0.74}
\usepackage[pagebackref,breaklinks,colorlinks,allcolors=cvprblue,bookmarks=false]{hyperref}

\makeatletter
\newif\ifappcites
\newcount\mainbibcount
\renewcommand\NAT@sort@cites[1]{%
  \let\NAT@cite@list\@empty
  \@for\@citeb:=#1\do{\expandafter\NAT@star@cite\@citeb\@@}%
  \if@filesw
    \ifappcites\else
      \expandafter\immediate\expandafter\write\expandafter\@auxout
        \expandafter{\expandafter\string\expandafter\citation\expandafter{\NAT@cite@list}}%
    \fi
  \fi
  \@ifnum{\NAT@sort>\z@}{%
    \expandafter\NAT@sort@cites@\expandafter{\NAT@cite@list}%
  }{}%
}
\newcommand{\beginappendixcites}{%
  \global\mainbibcount=\c@NAT@ctr
  \global\appcitestrue}
\newenvironment{appendixbibliography}{%
  \let\BR@starttoc\relax
  \let\GMDP@origbib\thebibliography
  \def\thebibliography##1{\GMDP@origbib{##1}\global\c@NAT@ctr\mainbibcount}}{}
\makeatother

\def\confName{CVPR}
\def\confYear{2027}

\title{General-purpose Medical Dense Prediction via Visual In-context Learning}

\author{
Yuntian Bo$^{1}$\quad
Piotr Koniusz$^{2,3}$\quad
Haofeng Zhang$^{1}$\\[1ex]
$^{1}$Nanjing University of Science and Technology\hspace{0.4em}
$^{2}$University of New South Wales\hspace{0.4em}
$^{3}$Data61$\!${\color{red}\heart}CSIRO\\[1ex]
{\normalsize\texttt{Project page:}\quad\href{https://primebo1.github.io/MedGen/}{\texttt{\color{magenta}https://primebo1.github.io/MedGen/}}}
}

\begin{document}
\maketitle

\begin{abstract}
  We explore how to make medical vision models generalizable across diverse tasks and scenarios by introducing a General-purpose Medical Dense Prediction (GMDP) task.
Inspired by recent advances in in-context learning (ICL) from NLP, we propose a Generalist Registration-Aware In-context Learner (GRAIL), which leverages the spatial similarity of paired medical images to infer targets through visual context prompts. A Spatial Registration Transformer (SRT) is introduced to model the semantic deformation between images, enabling the transformation of context labels into the target.
To establish a benchmark for GMDP evaluation, we also introduce Medical Generalist (MedGen), a comprehensive dataset covering five mainstream medical dense prediction tasks with over 77K labeled images, well-organized for developing GMDP algorithms.
Extensive experiments demonstrate that our approach consistently surpasses existing baselines by large margins.
\end{abstract}

\section{Introduction}

\begin{figure}[t]
    \centering
    \includegraphics[width=0.99\linewidth]{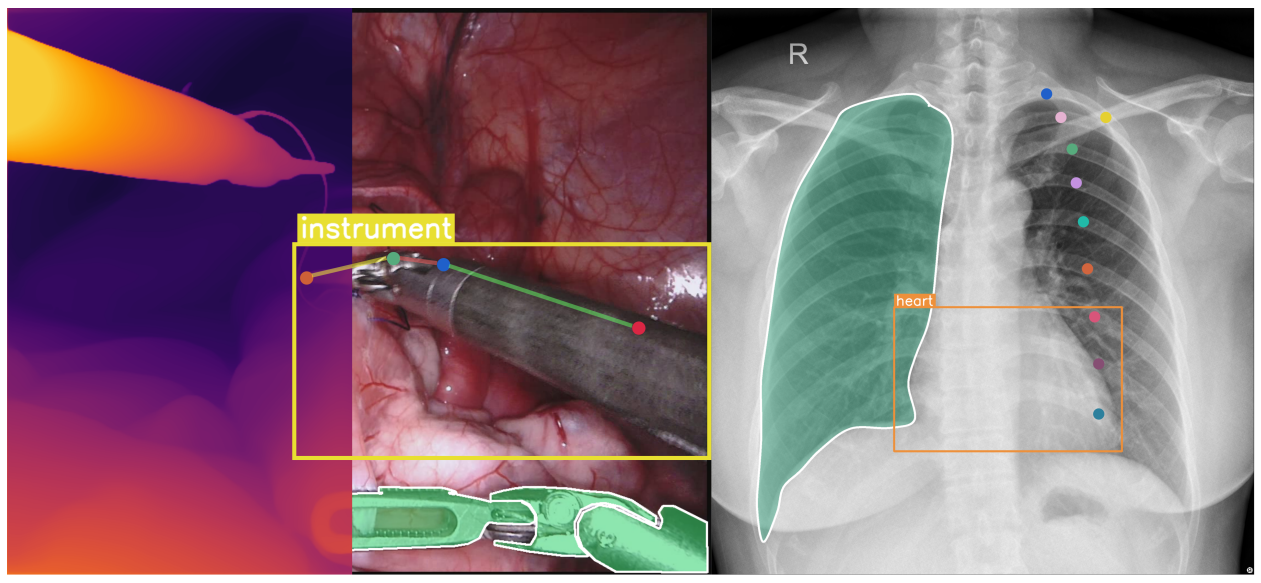}
    \caption{Medical visual analysis frequently involves multiple dense prediction tasks under different scenarios, such as (a) endoscopic surgery, which simultaneously requires depth estimation (DE), object detection (OD), segmentation (Seg), and landmark detection (LD); and (b) chest radiograph analysis, which involves Seg, OD, and LD. However, existing medical models remain limited to a single task.}
    \label{fig:motivation}
\end{figure}

The generality of visual models is a highly valuable property and a long-standing research goal, especially for real-world applications.
In the medical domain, computer-assisted visual analysis often requires multi-task collaboration to support diverse clinical activities, as illustrated in Figure \ref{fig:motivation}.
This, however, poses a significant generalization challenge for current medical vision models, which are typically designed for a single task or scenario.\footnote{ We refer to a \textit{task} as a visual understanding problem (\eg, segmentation or detection), and a \textit{scenario} as a combination of region of interest and imaging modality (\eg, \underline{abdominal} \underline{CT}).}

Recently, several studies have moved beyond conventional specialized paradigms toward building more general medical vision models, such as segmentation foundation models  UniverSeg \cite{UNIVERSEG} and MedSAM \cite{MEDSAM}.
While these few- or zero-shot methods address broader distribution of medical data, their focus remains limited and homogeneous, \ie, they remain task-specific in nature.
In contrast to natural image understanding, medical image analysis tasks are relatively more focused and structured, typically involving dense prediction problems (\ie, pixel-level understanding tasks), such as segmentation, object/landmark detection, and depth estimation, to support quantitative analysis and measurement.

\begin{figure*}[t]
    \centering
    \includegraphics[width=0.99\linewidth]{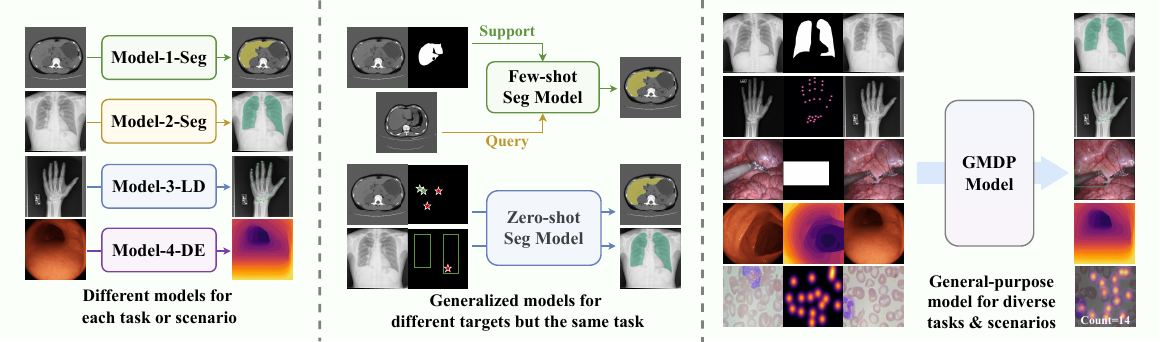}
    \caption{Task-specific models (left) require separate training/inference across medical scenarios and tasks. Few/zero-shot models (middle) generalize to new targets/scenes but remain single-task. While our GMDP framework (right) tackles diverse tasks across varied medical scenes with a unified model, without retraining or fine-tuning.}
    \label{fig:paradigm}
\end{figure*}

These insights naturally raise a question: \textit{Can we move beyond task-specific solutions and address multiple tasks within a single framework?}
Motivated by this, we formulate a novel task, termed General-purpose Medical Dense Prediction (GMDP), which aims to build unified models capable of handling diverse tasks across a wide range of medical scenarios, as illustrated in Figure \ref{fig:paradigm}.
However, considering the task–scenario dependency (\eg, depth estimation is feasible only in camera-based modalities such as endoscopy but not in radiological ones like MRI), determining what task and target to predict remains an open challenge for a model with a unified architecture for all tasks.

To address the aforementioned challenges, we draw inspiration from the recent success of In-context Learning (ICL) \cite{IN-CONTEXT_LEARNING} in NLP. This paradigm enables large language models to perform diverse and even unseen tasks without parameter tuning by taking task demonstrations as contextual examples in the input. Similarly in the visual domain, ICL can be adapted by using image–label pairs as visual prompts \cite{MAEINPAINTING}, which specify the task-required input–output mapping for in-context inference \cite{PAINTER,SD_VICL}.
This makes it possible to build a unified architecture without explicit task-type indication or task-specific structures.

Motivated by the above discussion, we propose a novel GMDP framework termed Generalist Registration-Aware In-context Learner (GRAIL).
Given a context pair (a reference image and its corresponding label), GRAIL reasons the input–output mapping to generate predictions for the target image in context, which makes GRAIL architecturally unified yet versatile for various tasks.
Specifically, GRAIL leverages the inherent spatial similarity between medical image pairs from the same scenario, and performs token-wise registration to transfer contextual information for target prediction. As its core, a Spatial Registration Transformer (SRT) is proposed, which learns precise latent deformations through adaptive spatial perception and implicit semantic registration. By learning to model context–target correspondence, GRAIL generalizes across tasks and scenarios, effectively mitigating the adverse effects of the huge variations in medical tasks and scenarios.

To standardize GMDP evaluation and pave the way for future development, we introduce a large-scale benchmark dataset, termed Medical Generalist (MedGen). Derived and re-organized from existing datasets, MedGen encompasses five mainstream dense prediction tasks across six common medical imaging modalities. The dataset contains over 77K images and more than 101K image–label pairs, covering a wide spectrum of medical analysis scenarios.
We benchmark GMDP using MedGen by assessing both in-distribution accuracy and out-of-distribution generalization. Extensive experiments on MedGen demonstrate that unifying medical visual tasks is not only feasible, but also enables individual tasks to benefit from a shared and transferable understanding of medical scenes beyond isolated task-specific learning.

The main contributions of our work are threefold:
\begin{enumerate}[label=\Roman*., leftmargin=*]
    \item We formulate the task of General-purpose Medical Dense Prediction (GMDP), which aims to build generalist models capable of solving diverse dense prediction tasks across a wide range of scenarios for medical visual analysis.
    \vspace{2pt}

    \item We approach GMDP through visual in-context learning, and propose a Generalist Registration-Aware In-context Learner (GRAIL). A Spatial Registration Transformer (SRT) is designed to boost registration-aware in-context inference by exploiting the spatial correspondence between paired medical images.
    \vspace{2pt}

     \item We establish a comprehensive GMDP benchmark by introducing the Medical Generalist (MedGen) dataset, which covers five mainstream tasks across multiple imaging modalities with over 77K labeled images, providing a data foundation for advancing research in this field.

\end{enumerate}

\section{Related Works}
\label{sec:related_works}
\noindent \textbf{Multi-task Learning (MTL).}
MTL aims to obtain multiple outputs associated with different tasks from a single input.
An intuitive way to achieve this is to attach multiple task-specific heads to a shared backbone \cite{UBERNET,Uni_PERCEPTUAL_PARSING,Cross-stitch,2024CVPRjointMTL}. Several approaches further improve multi-task synergy by introducing task balancing strategies to achieve collaborative optimization or to alleviate conflicts among tasks \cite{MTLUNCERTAINTY_LOSS,Gradnorm,DWA,DTP}.
However, conventional MTL models typically fit a set of predefined input–output mappings, leading to fixed outputs for specific scenes (\eg, indoor scenes).
In contrast, our paradigm employs a shared encoder–decoder and performs tasks entirely indicated by visual context prompts, enabling parameter-efficient and flexible dense prediction without overfitting to specific domains.

\noindent \textbf{Few-shot Learning (FSL). }
Our proposed paradigm is closely related to FSL, which similarly allows models to learn concepts from only a few examples.
A major branch of FSL focuses on optimization-based methods \cite{MAML,PROBMAML,OP_OD1,OP_OD2,OP_SEG}, which achieve rapid adaptation through a few gradient updates during inference, yet still rely on task-specific parameter fine-tuning.
Another dominant branch is metric learning-based methods \cite{PROTONET,siameseNET,RELATIONET,PANET,POMNET}, which learn discriminative feature spaces to directly label queries without further adaptation.
However, both paradigms are typically confined to single-task settings.
Recently, studies have attempted multi-task FSL.
For instance, VTM \cite{VTM} and UniAP \cite{UNIAP} design task-agnostic matching mechanisms for diverse dense prediction tasks, while UniFS \cite{UNIFS} treats point representations as unified descriptors for various instance perception tasks.
However, these models still regard test-time tuning as an essential step for adaptation.
In contrast, our model remains completely frozen during inference and can handle multiple dense vision tasks within a unified framework.

\noindent \textbf{In-context Learning (ICL).}
ICL was first explored in \cite{IN-CONTEXT_LEARNING}, referring to the ability of pretrained language models to adapt to or recognize a desired task from only a few prompts, without any parameter fine-tuning.
In the visual domain, several pioneering works \cite{VICL_GLOBAL_OP,VICL4LVLM,LZC_VICL,ECCV_VICL,GOOD_EXAMPLE_VICL,SINE,MMICL} extended this paradigm to build generalist models that perform similar in-context inference through visual prompt pairs.
For instance, Bar et al.~\cite{MAEINPAINTING} and Wang et al.~\cite{PAINTER} formulated visual ICL as an image in-painting problem~\cite{IMAGE_INPAINTING}, where the context image, its label, and the target image are concatenated as parts of an image, enabling the model to reconstruct the missing region corresponding to the target label.
SegGPT~\cite{SEGGPT} extended this idea by replacing in-painting with random color mapping of segmented regions. SD-VICL~\cite{SD_VICL} proposed a training-free approach by re-computing the intermediate attention of the Stable Diffusion ~\cite{STABLE_DIFFUSETION} to achieve context-guided target inference.

However, we observe that these methods perform poorly on medical images, which we attribute to their reliance on large-scale data for semantic modeling, whereas medical datasets are often too small. Existing medical approaches \cite{Show_and_Segment,SEGMIC,Tyche} attempt to adopt ICL, yet they are primarily tailored to \textit{segmentation} and lack generality across diverse task formulations. In contrast, our model focuses more on spatial correspondence modeling within context–target pairs through token-wise deformation, thereby reducing dependence on semantic understanding while enhancing generalization to diverse scenarios and tasks.

\section{MedGen Benchmark}

\subsection{Composition and Properties}
\begin{figure*}[t]
    \centering

    \begin{minipage}[t]{0.52\linewidth}
        \centering
        \begin{overpic}[width=\linewidth]{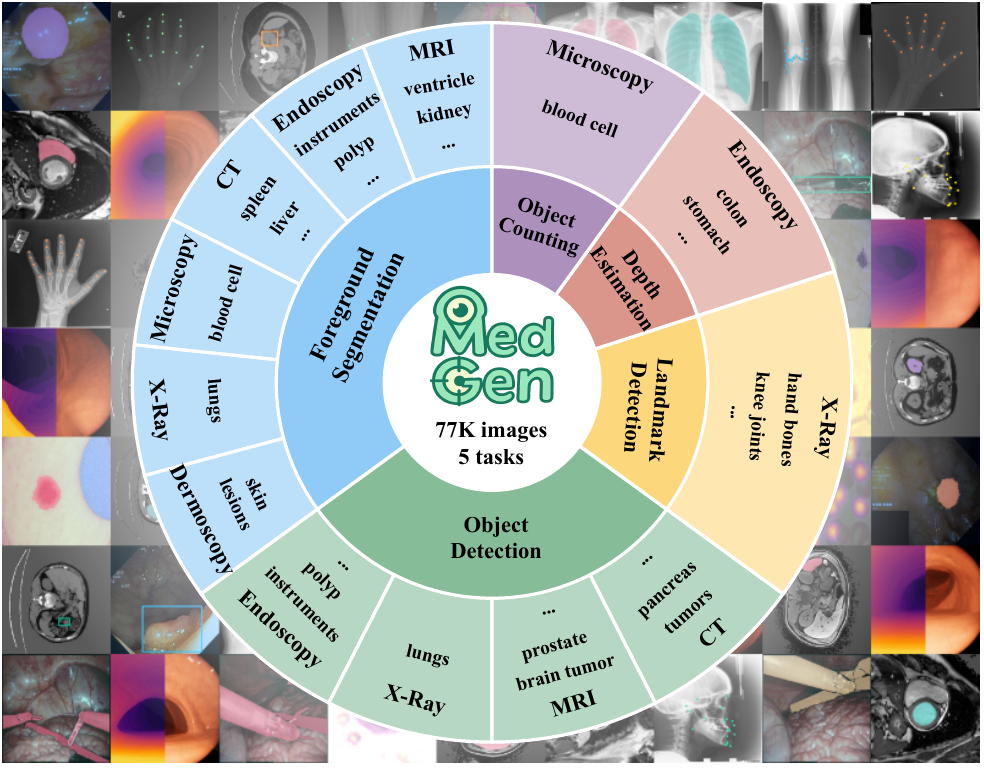}
            \put(0,79){\textbf{(a)}}
        \end{overpic}
    \end{minipage}
    \hfill
    \begin{minipage}[t]{0.46\linewidth}
        \centering
        \begin{overpic}[width=\linewidth]{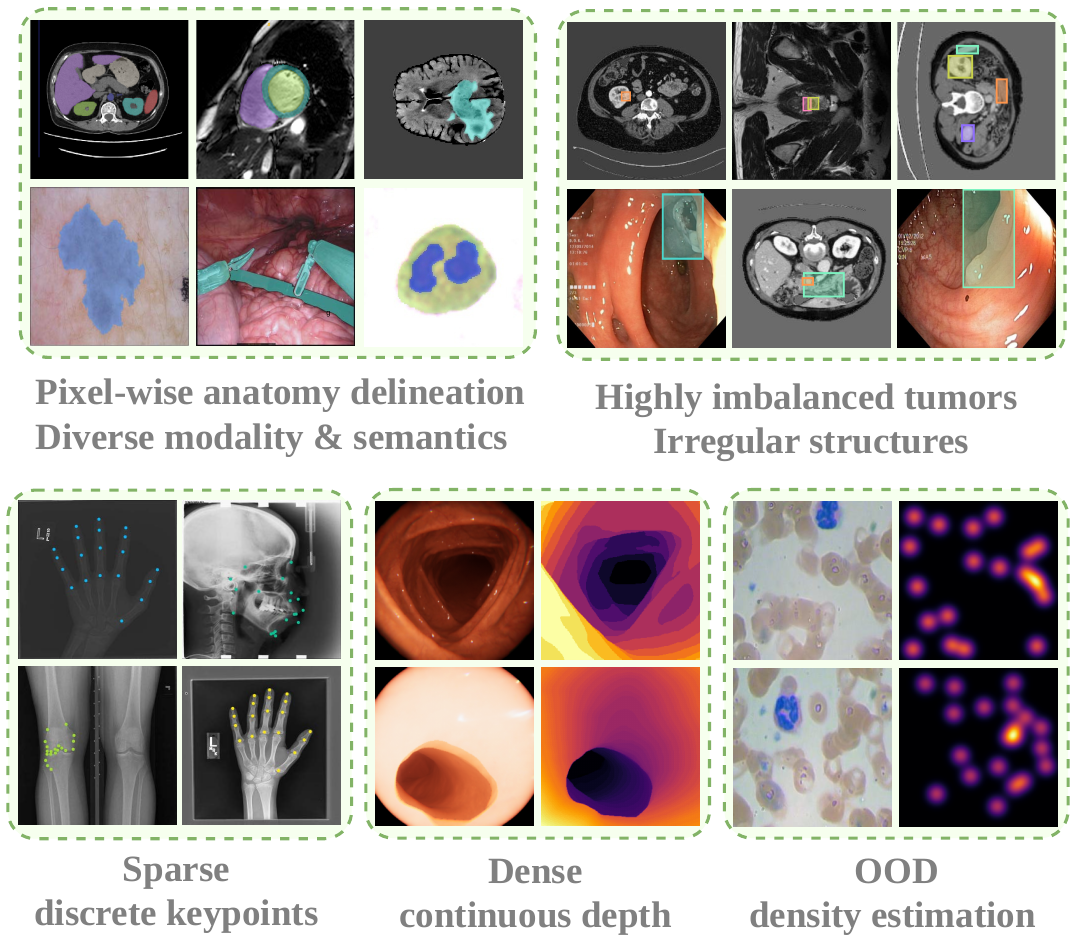}
            \put(0,89){\textbf{(b)}}
        \end{overpic}
    \end{minipage}
    \caption{\textbf{Overview of MedGen benchmark.} \textbf{(a) Visualization of MedGen's composition and diversity.} MedGen stands out for its broad coverage of medical visual analysis tasks, targeting over 20 anatomical structures and clinical entities across six common imaging modalities. A detailed description of each dataset is provided in Appendix~\ref{sec:medgen_detail}. \textbf{(b) Highly heterogeneous task focuses in MedGen.} This heterogeneity provides a challenging benchmark for evaluating models' unified understanding and handling of complex outputs across multiple visual analysis tasks.}
    \label{fig:MedGen}
\end{figure*}
Previous medical imaging datasets typically focus on a single task or combine several tasks within a restricted anatomical region, leaving no existing benchmark for evaluating GMDP. We therefore introduce Medical Generalist (MedGen), a large-scale, multi-task, and multi-scenario benchmark comprising over 77K medical images and 101K image--label task pairs. As illustrated in Figure~\ref{fig:MedGen}, MedGen is characterized by three key properties:

\begin{enumerate}[label=\Roman*., leftmargin=*]
\item \textbf{Broad task coverage.}
MedGen encompasses five structurally distinct dense prediction tasks: segmentation (Seg), object detection (OD), landmark detection (LD), depth estimation (DE), and object counting (OC). Their labels range from pixel-level masks and dense depth maps to bounding boxes, sparse keypoints, and discrete counts, presenting substantially different prediction objectives within a single benchmark.

\item \textbf{Diverse medical scenarios.}
MedGen spans six common imaging modalities, including X-ray, CT, MRI, endoscopy, dermoscopy, and microscopy, and covers more than 20 medical entities, such as organs,  lesions, blood cells, and surgical instruments. We also include extremely imbalanced or highly variable targets (\eg, tumors and the pancreas), which typically fall beyond the capabilities of generalist models, to provide a particularly challenging evaluation setting.

\item \textbf{Harmonized task formulation.}
Despite the substantial differences among task annotation formats, we carefully process and standardize all tasks into a consistent image-label interface. This unified formulation converts heterogeneous medical tasks into a common input-output mapping for dense prediction, making MedGen easy to use and enabling effective joint learning across tasks.

\end{enumerate}

We highlight that MedGen is particularly challenging due to the pronounced heterogeneity in predictive focus across its constituent tasks, as shown in Figure \ref{fig:MedGen} (b). Further task descriptions and an analysis of their differing focuses are provided in Appendix \ref{sec:appendix_Task_Descriptions}.

\begin{discussionbox}
\noindent \textbf{Motivation for Task Coverage.}
MedGen incorporates diverse tasks commonly encountered in real clinical activities. Specifically, OD provides coarse localization to constrain the search space for subsequent fine-grained analysis of focal regions. DE enables spatial understanding and supports surgical planning. LD facilitates precise measurement, tracking, and registration. Seg further delineates a wide range of structures and lesions commonly encountered in clinical practice, supporting quantitative diagnosis and interpretation.
The aimed all-in-one algorithm can be seamlessly integrated in practical clinical workflows, \eg, in surgical scenarios, DE and OD facilitate spatial understanding and ROI localization, while LD and Seg enable quantitative analysis and tool or tissue tracking.
\end{discussionbox}

\noindent \textbf{Format.}
Due to the wide coverage of imaging modalities, the original data are stored in diverse formats. We standardize them into a unified representation. For 2D images, we resize each image to $256 \times 256$ and normalize pixel intensities to the range $ [0,255] $. For 3D scans, we resample to a uniform voxel spacing, apply a center crop of $256 \times 256$, and then extract 2D slices containing targets, which are subsequently normalized to $ [0,255] $.
For multi-class Seg/OD datasets, we decompose each label map into multiple binary masks of each class and store them separately. Before model input, all 2D slices are replicated along the channel dimension to three channels if grayscale, ensuring consistency with RGB image formats.

\subsection{Evaluation Protocol}
\noindent \textbf{Evaluation Scenarios.}
For clinical applications, methods should demonstrate strong performance on scenarios seen during training as well as robustness to previously unseen scenarios. We benchmark two evaluation settings:
1)	In-distribution (ID) scenario. The model is evaluated on data drawn from the same distribution and vision tasks as in training, including OD, DE, Seg, and LD.
2) Out-of-distribution (OOD) scenario.
We evaluate three types of distribution shifts: domain shift (\eg, training on AMOS-CT \cite{AMOS} but testing on AMOS-MRI \cite{AMOS}) and task shift (\eg, Chest X-ray \cite{Chestxray1,Chestxray2} used for OD during training but for Seg during testing). In addition, we further consider OC to assess the model’s inference-time generalization to a task that is entirely unseen during training.

\noindent \textbf{Data Split.}
To comprehensively assess a generalist model, MedGen is first partitioned into an ID  set and an OOD set. Each dataset in the ID set is then randomly split into training (60\%), validation (20\%), and test (20\%) subsets. For 3D volumes, the split is performed at the case level before converting them into 2D slices, ensuring that data from the same patient never appears in both training and test sets. Datasets in the OOD set are kept intact and used only for testing. We provide dataset usage details in Appendix \ref{sec:medgen_detail}.

\noindent \textbf{Metrics.}
Following the standard conventions of each task and previous VICL methods \cite{PAINTER,MAEINPAINTING}, we adopt Normalized Mean Squared Error (NMSE) for DE, Dice Similarity Coefficient (DSC) and 95th percentile Hausdorff Distance (HD95) for Seg, mean Intersection-over-Union (mIoU) for OD, Percentage of Correct Keypoints at 0.1 of the image size (PCK@0.1) and NMSE for LD, and Mean Absolute Error (MAE) and Root Mean Square Error (RMSE) for OC. The detailed definitions of each metric are provided in Appendix \ref{sec_evaluation_metric}.

\begin{figure*}[t]
    \centering
    \includegraphics[width=0.99\linewidth]{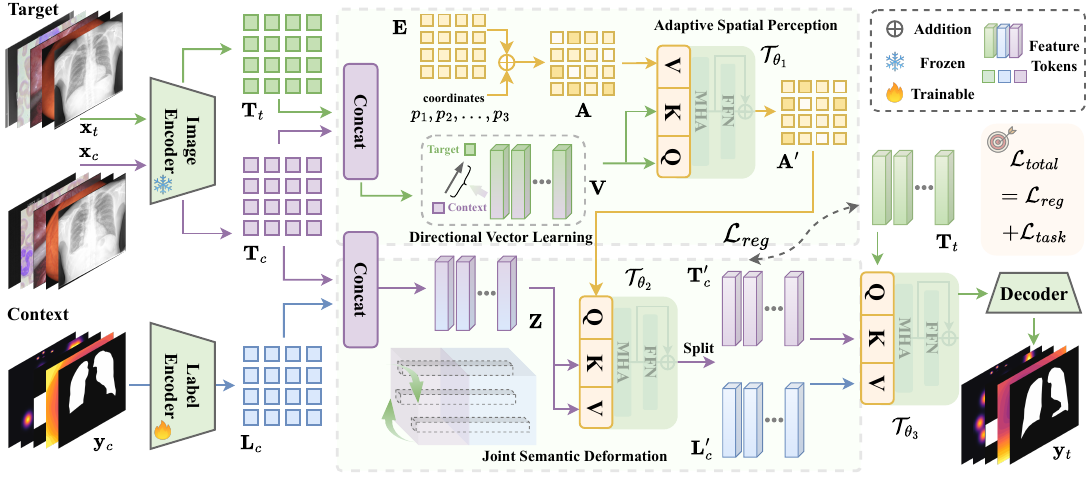}
    \caption{Overview of Our Proposed GRAIL. Image and label encoders extract the token representations of the input images and the context label, respectively. \SRTbadge learns to register the context tokens to the target tokens by adaptively modeling deformation directions and implicit context–target registration, which facilitates subsequent correspondence-based decoding.}
    \label{fig:overall_arch}
\end{figure*}

\section{Methodology}

\subsection{Problem Definition}
In the framework of GMDP via ICL, a unified model $\Theta$ is designed to handle diverse dense prediction tasks $\mathcal{T}$.
The task type is implicitly indicated by a context image–label pair $\mathcal{C} = (\mathbf{x}_c, \mathbf{y}_c)$,
which consists of a context image $\mathbf{x}_c \in \mathbb{R}^{3 \times H \times W}$ and its corresponding label $\mathbf{y}_c \in \mathbb{R}^{H \times W}$,
defining the task-specific input–output mapping.
By treating $\mathcal{C}$ as a visual prompt, $\Theta$ predicts the label map of the target image $\mathbf{x}_t$ for task $\mathcal{T}$.

The model $\Theta$ is trained on a large-scale dataset $\mathcal{D}$ containing diverse densely annotated image-label pairs across multiple tasks.
In each iteration, two pairs from the same task are sampled, one serving as the context and the other as the target, enabling the model to infer the target conditioned on the input–output mapping specified by the context. During inference, model parameters are frozen without any update. The evaluation covers multiple tasks under both in-distribution and out-of-distribution settings \textit{w.r.t.} $\mathcal{D}$.

\subsection{Method Overview}
The overview of our proposed GRAIL framework is illustrated in Figure \ref{fig:overall_arch}.
To enable effective context–target interactions, GRAIL is built upon a Transformer \cite{TRANSFORMER} architecture with four major components: a frozen pretrained backbone, a label backbone, a Spatial Registration Transformer (SRT), and a lightweight context decoder.

\subsection{Feature Extraction}
We employ two parallel feature extractors to obtain tokens for images and labels, respectively.
In our implementation, the frozen image encoder $f_{I}(\cdot)$ is adopted from the DINOv3 family \cite{DINOV3}, while the label encoder $f_{L}(\cdot)$ is an ImageNet\cite{imagenet}-pretrained ViT-B \cite{VIT}.
Through this design, for arbitrary context–target triplet
$(\mathbf{x}_c, \mathbf{y}_c, \mathbf{x}_t)$,
we obtain the corresponding feature tokens as follows:
the context visual tokens
$\mathbf{T}_c = f_{I}(\mathbf{x}_c)$,
the context label tokens
$\mathbf{L}_c = f_{L}(\mathbf{y}_c)$,
and the target visual tokens
$\mathbf{T}_t = f_{I}(\mathbf{x}_t)$.
All tokens share the same dimension of
$\mathbb{R}^{N \times C}$, where $N = \frac{HW}{\mathrm{p}^2} $ denotes the number of patch tokens, $\mathrm{p}$ denotes the patch size, and $C$ denotes the embedding dimension.

\subsection{Spatial Registration Transformer}
The design of SRT is motivated by the inherent spatial similarity observed in medical image pairs. Despite individual anatomical variations that cause local displacements, medical images from the same scenario generally share highly similar spatial structures, allowing spatial–semantic alignment to be achieved through minor deformations. Our key insight is that in-context prediction can be reformulated as learning a latent deformation that spatially registers the context image tokens to those of the target. The same deformation can be applied to the context label tokens to obtain the target ones for decoding the target label. We formulate SRT as a two-stage architecture composed of Adaptive Spatial Perception (ASP) and Joint Semantic Deformation (JSD).

\subsubsection{Adaptive Spatial Perception}
\label{sec:ASP}
In this stage, we aim to perceive deformation information and establish a latent carrier to guide the deformation of the context toward the target.
Notably, under the multi-scenario nature of the GMDP formulation, the spatial properties of input images are uncertain and may differ substantially on an instance-wise basis.
However, despite scenario-level variations, the semantic displacement between corresponding spatial regions of the context and target is inherently small. This motivates us to examine the differential relationships between regions, represented by image patch tokens, to adaptively reason deformation directions.

To this end, firstly, we propose a directional vector formulation, where each vector is constructed by taking the context and target image tokens as its start and end points, respectively, to infer their deformation displacement:
\begin{equation}
\mathbf{V}(i)= \xi \big(\operatorname{concat}(\mathbf{T}_c(i), \mathbf{T}_t(i))\big) \in \mathbb{R}^{2C},  i\in\{1,\dots,N\},
\end{equation}
where $\mathbf{V} \in \mathbb{R}^{N \times 2C}$ denotes the directional vectors, $\xi : \mathbb{R}^{2C} \to \mathbb{R}^{2C}$ is an MLP for vector reasoning,
 $\operatorname{concat}(\cdot, \cdot)$ denotes a channel-wise concatenation function, $\mathbf{T}(i) \in \mathbb{R}^{C}$ denotes the feature token at the $i$-th sequence position.

Secondly, we utilize the spatial displacement encoded in $\mathbf{V}$ to adaptively generate a deformation anchor $\mathbf{A}$, which is later used as a carrier to propagate the reasoned deformation onto the context tokens.
The anchor $\mathbf{A}$ is initialized as a set of spatial tokens, formulated as:
\begin{equation}
\mathbf{A}  = \mathbf{E} + \Gamma(\mathbf{P}) \in \mathbb{R}^{N \times 2C},
\end{equation}
where $\mathbf{E} \in \mathbb{R}^{N \times 2C}$ denotes a set of learnable tokens, and $\Gamma(\mathbf{P})$ denotes the positional encoding that serves as the coordinate basis for deformation modeling, transformed from the spatial coordinates $\mathbf{P}=\{p_i\}_{i=1}^N$ following \cite{fourier_PE}:
\begin{equation}
\Gamma(\mathbf{P})
=
\operatorname{concat}\big(
\cos\big(2\pi\,\mathbf{P}\mathbf{B}^{\top}\big)
 , \;
\sin\big(2\pi\,\mathbf{P}\mathbf{B}^{\top}\big)
\big)
\in \mathbb{R}^{N \times 2C},
\end{equation}
where $\mathbf{P}\in\mathbb{R}^{N\times 2}$ encodes each token's normalized 2D coordinates in $[-1,1]$,
$\mathbf{B}\in\mathbb{R}^{C\times 2}$ has entries sampled from a Gaussian distribution $\mathcal{N}(0,\sigma^{2})$.
We propagate information from $\mathbf{V}$ to $\mathbf{A}$ through a cross-attention Transformer,
enabling $\mathbf{A}$ to instance-adaptively derive a direction-aware deformation anchor. This process is formulated as:
\begin{equation}
     \mathbf{A}' = \mathcal{T}_{\theta_1}(\mathbf{V}, \mathbf{V}, \mathbf{A}) + \mathbf{A} \in \mathbb{R}^{N \times 2C},
\end{equation}
where the anchor $\mathbf{A}'$ globally aggregates deformation directions.
$\mathcal{T}_{\theta_1}(\mathbf{q}, \mathbf{k}, \mathbf{v})$ denotes a Transformer parameterized by $\theta_1$, which takes the query $\mathbf{q}$, key $\mathbf{k}$, and value $\mathbf{v}$ as inputs according to their designated positions.

\subsubsection{Joint Semantic Deformation}
In this stage, we exploit the spatial deformation from the context image tokens $\mathbf{T}_c$ to the target image tokens $\mathbf{T}_t$
to guide the transformation of the context label tokens $\mathbf{L}_c$ toward the target label embedding space.
Since $\mathbf{T}_c$ and $\mathbf{L}_c$ share pixel-wise semantic correspondence for the same sample,
we naturally concatenate them within a unified coordinate system for joint deformation learning:
\begin{equation}
    \mathbf{Z} = \operatorname{concat}(\mathbf{T}_c, \mathbf{L}_c) \in \mathbb{R}^{N \times 2C},
\end{equation}
where $\mathbf{Z}$ denotes the concatenated tokens.
Inspired by image registration, we learn the deformation process for the aforementioned transformation.
Unlike traditional registration approaches that explicitly estimate a deformation field to warp the source into the target space,
our method implicitly models this process through a Transformer conditioned on the anchor $\mathbf{A}'$:
\begin{equation}
    \mathbf{Z}' = \mathcal{T}_{\theta_2} (\mathbf{A}',\mathbf{Z}, \mathbf{Z}) + \mathbf{Z} \in \mathbb{R}^{N \times 2C},
\end{equation}
where $\mathbf{Z}'$ denotes the deformed tokens.

To ensure that the learned deformation is optimized toward the correct direction,
we enforce spatial and semantic consistency between the context and target image tokens.
To this end, we first re-split $\mathbf{Z}'$ to obtain the deformed context tokens $\mathbf{T}_c'$ and $\mathbf{L}_c'$,
and then compute SSIM \cite{SSIM} and token-wise cosine similarity losses between $\mathbf{T}_c'$ and the target image tokens $\mathbf{T}_t$.
This process is formalized as:
\begin{equation}
\mathbf{T}_c', \mathbf{L}_c' = \operatorname{split}(\mathbf{Z}'; C),
\end{equation}
\begin{equation}
\begin{aligned}
\mathcal{L}_{reg}  &=
\mathcal{L}_{cos}(\mathbf{T}_c', \mathbf{T}_t)
+ \mathcal{L}_{ssim}(\mathbf{T}_c', \mathbf{T}_t) \\
     &= 1 - \operatorname{cos}(\mathbf{T}_c', \mathbf{T}_t) + 1 -  \operatorname{SSIM}(\mathbf{T}_c', \mathbf{T}_t),
\end{aligned}
\end{equation}
where $\operatorname{split}(\mathbf{t}; C):
\mathbb{R}^{ N\times 2C}
\to
(\mathbb{R}^{ N\times C})^2$
denotes a channel-wise split function that divides
$\mathbf{t}$ into two $C$-channel parts and $\mathcal{L}_{reg}$ denotes the combined loss for semantic registration. Here, $\mathcal{L}_{cos}$ focuses on enforcing semantic consistency,
and $\mathcal{L}_{ssim}$ emphasizes spatial structural consistency.

\begin{table*}[t]
\centering
\setlength{\tabcolsep}{4pt}
\renewcommand{\arraystretch}{1.2}

\caption{Comparison across multiple ID dense prediction tasks on MedGen. $\downarrow$ indicates lower is better, $\uparrow$ indicates higher is better. Best results are indicated in bold font, and second bests are underlined.}
\vspace{-2ex}
\begin{adjustbox}{max width=\textwidth}
\label{tab:main}
\begin{tabular}{lccccccc}
\toprule
\multirow{2}{*}{\textbf{Method}} &
\multicolumn{1}{c}{\textbf{Depth Estimation}} &
\multicolumn{2}{c}{\textbf{Foreground Segmentation}} &
\multicolumn{1}{c}{\textbf{Object Detection}} &
\multicolumn{2}{c}{\textbf{Landmark Detection}} \\
\cmidrule(lr){2-2}
\cmidrule(lr){3-4}
\cmidrule(lr){5-5}
\cmidrule(lr){6-7}
& NMSE$\downarrow$ &
HD95$\downarrow$ &
DSC$\uparrow$ &
mIoU$\uparrow$ &
NMSE$\downarrow$ &
PCK@0.1$\uparrow$ \\
\midrule
VTM \cite{VTM} & \underline{0.022} & \underline{36.48} & \underline{70.70} & 45.00 & 0.0236 & 79.92 \\
UniAP \cite{UNIAP} & 0.040 & 44.80 & 65.18  & 42.54 & 0.0271 & 72.54 \\
MAE-VQGAN \cite{MAEINPAINTING} & 0.031 & 40.84 & 66.06 & 44.75 & 0.0291 & 68.28 \\
Painter \cite{PAINTER} & 0.029 & 43.48 & 69.09 & 42.59 & 0.0237 & 70.88 \\

prompt-SeIF \cite{LZC_VICL} & 0.025 & 45.76 & 69.24 & \underline{44.87} & 0.0276 & 73.00 \\
 UniverSeg \cite{UNIVERSEG} & 0.165 &  51.63& 50.89 & 37.26 & 0.0310 & 65.49 \\
 Tyche \cite{Tyche} & 0.145 & 55.65 & 61.32 & 40.95 & \underline{0.0185} & \underline{80.96} \\
 SegMIC \cite{SEGMIC} & 0.041 & 55.63 & 66.12 &  42.44 &  0.0297 & 71.03 \\
 \rowcolor{blue!8}
\textbf{GRAIL (Ours)} & \textbf{0.019} & \textbf{35.23} & \textbf{72.40} & \textbf{46.18} & \textbf{0.0137} & \textbf{82.19} \\
\bottomrule
\end{tabular}
\end{adjustbox}
\end{table*}

\subsection{Decoding by Correspondence}
We measure the semantic correspondence between the deformed context image tokens $\mathbf{T}_c'$ and the target tokens $\mathbf{T}_t$, and use the deformed label tokens $\mathbf{L}_c'$ to gate and project this correspondence into the target output space. This is implemented via a Transformer where $\mathbf{T}_c'$ serves as the query, $\mathbf{T}_t$ as the key, and $\mathbf{L}_c'$ as the value. Subsequently, a lightweight decoder is employed to produce the final prediction $\tilde{\mathbf{y}_t}$:
\begin{equation}
    \tilde{\mathbf{y}_t} = \phi \big(\mathcal{T}_{\theta_3}(\mathbf{T}_t,\mathbf{T}_c', \mathbf{L}_c')\big) \in \mathbb{R}^{1 \times H \times W},
\end{equation}
where
$\phi(\cdot): \mathbb{R}^{C \times \frac{HW}{\mathrm{p}^2}} \to \mathbb{R}^{1 \times H \times W}$
denotes the decoder composed of a reshape function, transposed convolutions, and ReLU activation that upscales the spatial resolution by a factor of $p$.
We posit that stronger registration in SRT yields more accurate context–target correspondence responses, thereby enhancing in-context inference.

The total training loss of GRAIL is formulated as
$\mathcal{L}_{total} = \mathcal{L}_{reg} + \mathcal{L}_{task}(\tilde{\mathbf{y}_t}, \mathbf{y}_t),$ where the task loss $\mathcal{L}_{task}$ is instantiated as the cross-entropy loss for Seg and OD, and as the MSE loss for LD and DE.

\section{Experiments}

\subsection{Implementation Details}
We implement our proposed method in PyTorch \cite{pytorch}. Unless otherwise specified, all experiments are conducted on NVIDIA Tesla V100 32GB GPUs. We train each model for 500K iterations using the Adam optimizer \cite{adam} with a learning rate of $1 \times 10^{-5}$ and a batch size of 1 on a single GPU. Validation is performed after every 10K iterations and we select the checkpoint with the best validation performance for testing. To mitigate optimization conflicts among different tasks, we adopt GradNorm \cite{Gradnorm} to balance the gradient magnitudes. We employed DINOv3 \cite{DINOV3} `ViT-B/16' version as the image encoder $f_{I}(\cdot)$ in our experiments. During training, $f_{I}(\cdot)$ is kept frozen, while the label encoder $f_{L}(\cdot)$ is fine-tuned.

\subsection{Benchmark Results on MedGen}
\label{sec:Benchmark_result}
In Table~\ref{tab:main}, we report the benchmark results of our proposed GRAIL and other open-sourced \footnote{ Some methods mentioned in Section \ref{sec:related_works} are not included in the comparison due to the lack of publicly available code or incompatible data interfaces (see Appendix \ref{sec:baseline_coverage}).} baselines on MedGen.
The compared baselines include the universal FSL methods VTM \cite{VTM} and UniAP \cite{UNIAP}, MAE-based ICL methods MAE-VQGAN \cite{MAEINPAINTING}, Painter \cite{PAINTER} and prompt-SeIF (+MAE-VQGAN) \cite{LZC_VICL}, and adapted medical in-context \textit{segmentation} methods UniverSeg \cite{UNIVERSEG}, Tyche \cite{Tyche} and SegMIC \cite{SEGMIC}.

\subsubsection{In-distribution Evaluation}
Under the ID setting, we evaluate the models’ performance on scenarios and tasks that are seen during training,  as shown in Table \ref{tab:main}. Detailed results for each dataset are provided in Appendix \ref{sec:Results_on_Individual_Datasets}.

\noindent \textbf{Foreground Segmentation.}
Our method achieves the best DSC of 72.40\% among all baselines. Notably, it also yields a substantially lower HD95 (by an average of 6.93) compared with MAE-based methods \cite{MAEINPAINTING,PAINTER}, indicating more accurate boundary alignment achieved through spatial registration rather than mere semantic fitting.
We further observe that our method performs particularly well on datasets where each image contains multiple targets.

\begin{table}[t]
\centering
\setlength{\tabcolsep}{3pt}
\renewcommand{\arraystretch}{1.15}
\caption{Comparison under out-of-distribution settings.}
\label{tab:OOD}
\vspace{-2ex}
\resizebox{\linewidth}{!}{
\begin{tabular}{lcccc}
\toprule
\multirow{2}{*}{\textbf{Method}}
& \multicolumn{1}{c}{\textbf{Domain Shift}}
& \multicolumn{1}{c}{\textbf{Task Shift}}
& \multicolumn{2}{c}{\textbf{Unseen Task}} \\
\cmidrule(lr){2-2}\cmidrule(lr){3-3}\cmidrule(lr){4-5}
& DSC $\uparrow$ & DSC $\uparrow$ & MAE $\downarrow$ & RMSE $\downarrow$ \\
\midrule
VTM \cite{VTM} & 45.19 & 79.26 & 7.80& 7.92\\
UniAP \cite{UNIAP} & 34.30 & 78.35 & 9.46 & 11.36 \\
MAE-VQGAN \cite{MAEINPAINTING} & 43.27 & 79.01 & 7.37 & 8.06 \\
Painter \cite{PAINTER} & 46.14 & \textbf{83.15} & \underline{6.79} & \underline{7.62} \\
prompt-SeIF \cite{LZC_VICL} & \underline{47.15} & 80.75 & 8.58 & 9.47 \\
UniverSeg \cite{UNIVERSEG} & 38.34 & 79.49 & 8.53 & 9.59\\
Tyche \cite{Tyche} & 28.19 & 80.32 & 10.40 & 11.19 \\
SegMIC \cite{SEGMIC} & 45.12 & 78.24 & 9.89 & 10.11 \\
\rowcolor{blue!8}
\textbf{GRAIL (Ours)} & \textbf{48.84} & \underline{81.86} & \textbf{6.12} & \textbf{7.37} \\
\bottomrule
\end{tabular}
}
\end{table}

\begin{figure*}[t]
    \centering
    \includegraphics[width=0.95\linewidth]{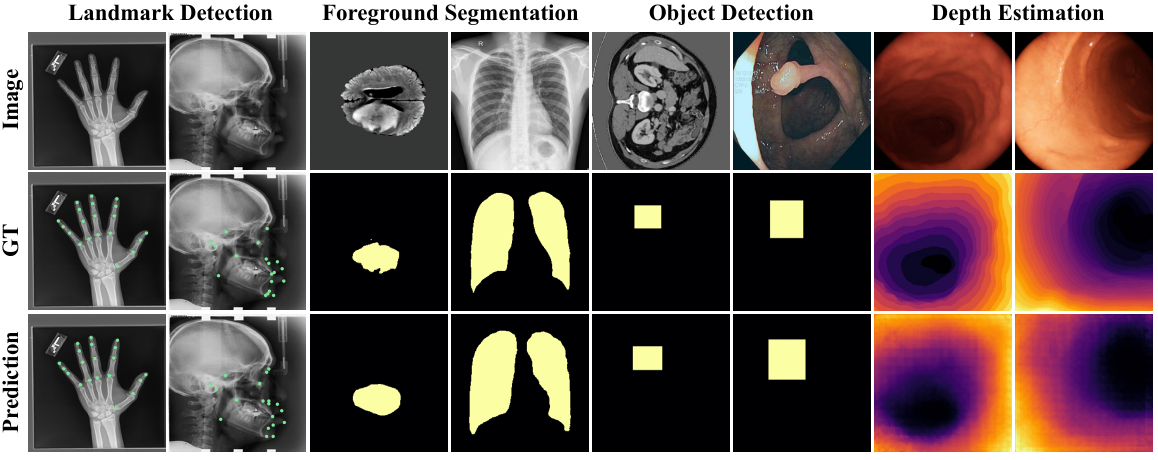}
    \vspace{-1ex}
    \caption{Visualization of example predictions generated by GRAIL across diverse tasks.}
    \label{fig:visualization}
\end{figure*}

\begin{table*}[t]
\centering
\setlength{\tabcolsep}{4pt}
\renewcommand{\arraystretch}{1.2}
\caption{Comparison of model performance between multi-task joint training and single-task training.}
\vspace{-2ex}
\label{tab:separate_training}
\begin{adjustbox}{max width=\textwidth}
\begin{tabular}{lcccccc}
\toprule
\multirow{2}{*}{\textbf{Model}} &
\multicolumn{1}{c}{\textbf{Depth Estimation}} &
\multicolumn{2}{c}{\textbf{Foreground Segmentation}} &
\multicolumn{1}{c}{\textbf{Object Detection}} &
\multicolumn{2}{c}{\textbf{Landmark Detection}} \\
\cmidrule(lr){2-2}
\cmidrule(lr){3-4}
\cmidrule(lr){5-5}
\cmidrule(lr){6-7}
& NMSE$\downarrow$ &
HD95$\downarrow$ &
DSC$\uparrow$ &
mIoU$\uparrow$ &
NMSE$\downarrow$ &
PCK@0.1$\uparrow$ \\
\midrule
Single-task training &
0.015 & 38.39 & 71.43 & 41.85 & 0.0251 & 70.48 \\
Multi-task training &
$0.019_{\textcolor{red}{ \scriptsize+0.004}}$ &
$35.23_{\textcolor{green!60!black}{ \scriptsize-3.16}}$ &
$72.40_{\textcolor{green!60!black}{ \scriptsize+0.97}}$ &
$46.18_{\textcolor{green!60!black}{ \scriptsize+4.33}}$ &
$0.0137_{\textcolor{green!60!black} {\scriptsize-0.0114}}$ &
$82.19_{\textcolor{green!60!black}{ \scriptsize+11.71}}$ \\
\bottomrule
\end{tabular}
\end{adjustbox}
\vspace{-1ex}
\end{table*}

\begin{table}[t]
\centering

\caption{Ablations on model components and their impact.}
\label{tab:module_effectiveness}
\vspace{-2ex}
\resizebox{0.95\linewidth}{!}{
\begin{tabular}{ccc>{\centering\arraybackslash}p{0.05\linewidth}ccccc}
\toprule
\multicolumn{3}{c}{SRT} & \multirow{2}{*}{$\mathcal{T}_{\theta_3}$} & \multicolumn{5}{c}{\textbf{Task}} \\
\cmidrule(lr){1-3} \cmidrule(lr){5-9}
$\Gamma(\mathbf{P})$ & $\mathbf{A}'$ & JSD & & DE$\downarrow$ & Seg$\uparrow$ & OD$\uparrow$ & LD$\uparrow$ & OC$\downarrow$ \\
\midrule
& & & \checkmark & 0.042 & 63.98 & 40.01 & 71.49 & --\\
& \checkmark & \checkmark & \checkmark & 0.031 & 70.51 & 44.12 & 76.82 & --\\
& & \checkmark & \checkmark & 0.036 & 69.61 & 41.23 & 75.66 & --\\
\checkmark & \checkmark & \checkmark & & 0.032 & 68.26 & 40.73 & 79.83 & 11.04\\
\checkmark & \checkmark & \checkmark & \checkmark & \textbf{0.019} & \textbf{72.40} & \textbf{46.18} & \textbf{82.19} & \textbf{7.37}\\
\bottomrule
\end{tabular}
}
\begin{tablenotes}
\item Here we ablate the coordinate basis $\Gamma(\mathbf{P})$ and whether the anchor $\mathbf{A}'$ in ASP is adaptively generated. Most variants retain JSD as it is essential for the registration stage. $\mathcal{T}_{\theta_3}$ without a ‘\checkmark’ indicates that the decoder directly uses the registered label tokens as input.
\end{tablenotes}
\end{table}

\noindent \textbf{Single Object Detection.}
Our method continues to achieve the best performance, yielding an mIoU of 46.18\%.
Compared with Seg, OD relies more on precise spatial localization, which further verifies the superiority of GRAIL in spatial perception.

\noindent \textbf{Landmark Detection.}
Our method also performs strongly on LD, surpassing the best competitor Tyche \cite{Tyche} by 1.23\% in PCK@0.1. However, although Tyche ranks second on LD, we observe that its performance on the other three tasks is unexpectedly weak (\eg, 11.08\% lower than ours in DSC, despite being originally designed for Seg). We attribute this discrepancy to its U-Net–like architecture, which struggles to generalize when substantial context–target discrepancies exist, a scenario that is common in MedGen. In contrast, our token-level deformation learning explicitly optimizes the mapping from context points to target point predictions, leading to consistently strong performance across all tasks.

\noindent \textbf{Depth Estimation.}
DE is the only task requiring full-pixel deformation with fully continuous predictions. Method emphasizing more on discrete space, \eg, MAE-VQGAN \cite{MAEINPAINTING} performs relatively worse, whereas our method achieves the best result with an NMSE of 0.019, indicating more precise per-pixel estimation.

\begin{table}[t]
\centering

\caption{Ablation studies on different loss formulations.}
\label{tab:loss_function}
\vspace{-2ex}
\resizebox{0.95\linewidth}{!}{
\begin{tabular}{lcccc}
\toprule
$\mathcal{L}_{reg}$ & DE$\downarrow$ & Seg$\uparrow$ & OD$\uparrow$ & LD$\uparrow$ \\
\midrule
$\mathcal{L}_{cos}$ only & 0.030 & 70.33 & 42.37 & 80.74\\
$\mathcal{L}_{ssim}$ only & 0.028 & 71.12 & 44.29 & 80.73\\
$\mathcal{L}_{cos} + \mathcal{L}_{ssim}$ & \textbf{0.019} & \textbf{72.40} & \textbf{46.18} & \textbf{82.19}\\
\bottomrule
\end{tabular}
}
\vspace{-7pt}
\end{table}

\subsubsection{Generalize Out-of-Distribution}
We evaluate model generalization under three settings: 1) \textit{Domain shift:} the model is trained on \textit{AMOS-CT} for Seg, but is evaluated on \textit{AMOS-MRI} for Seg.
2) \textit{Task shift}: the model is trained on Chest X-ray for \textit{OD}, yet is evaluated on Chest X-ray for \textit{Seg}.
3) \textit{Unseen task}: we directly assess the model on OC, which is completely held out during training.

\begin{figure*}[t]
    \centering
    \includegraphics[width=0.99\linewidth]{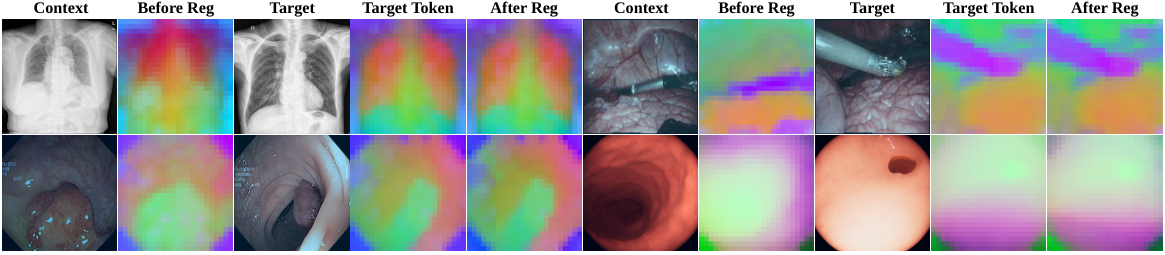}
    \vspace{-1ex}
    \caption{Visualization of feature comparisons for registration under large context–target variation. “Reg” denotes registration.}
    \label{fig:large_variation}
\end{figure*}

From Table~\ref{tab:OOD}, we conclude that our method exhibits the overall best generalization. Although Painter \cite{PAINTER} performs better under task shift, our model substantially outperforms it in the remaining settings. We attribute the superior generalization of GRAIL to its explicit modeling of context–target correspondence through spatial registration, rather than solely fitting on image semantics.

\subsection{Qualitative Analysis}
We present several qualitative results of GRAIL in Figure \ref{fig:visualization}, covering LD, Seg, OD and DE. As shown, our method also performs visually well, accurately predicting across diverse tasks and scenarios, demonstrating practical value in supporting clinical activities.

\subsection{Diagnostic Experiments}
\label{sec:ablations}

\noindent \textbf{Joint Training vs. Task-specific Training.} We compare joint training on MedGen with task-specific training on each task subset. As shown in Table~\ref{tab:separate_training}, unifying medical task handling has an overall positive effect, with two main findings. First, despite the high heterogeneity among tasks, most individual tasks, including Seg, OD, and LD, still benefit from joint training. In particular, tasks with limited data, such as LD, clearly benefit from the knowledge learned jointly with other tasks that have more data samples. Second, DE is the only task whose performance deteriorates under joint training. We hypothesize that this is because DE uniquely requires explicit 3D spatial reasoning due to its camera-based nature, whereas the other tasks are dominated by 2D scan-derived images. Consequently, training on these images may provide limited geometric supervision and may even introduce negative transfer. Future work could mitigate this conflict by incorporating dedicated 3D-aware dimensions into ViT tokens, following the design of VGGT~\cite{vggt}.

\noindent \textbf{Component Effectiveness.}
We ablate each component to examine its contribution, as shown in Table~\ref{tab:module_effectiveness}.
For Seg, LD, and OC, we report DSC, PCK@0.1, and RMSE as the performance measures.
SRT yields significant performance improvements (\eg, +8.42\% for Seg), mainly owing to its JSD component. The coordinate basis $\Gamma(\mathbf{P})$ is also indispensable for tasks with precise spatial information, such as LD.
Through the ablation of $\mathcal{T}_{\theta_3}$ (last two rows), we observe that correspondence-based decoding makes the model more adaptive and flexible across diverse scenarios and tasks, especially for unseen training tasks OC.

\noindent \textbf{Loss Function Formulation.}
We explore different loss variants in Table~\ref{tab:loss_function}. We find that both loss functions contribute significantly to learn better registration: $\mathcal{L}_{cos}$ focuses on maintaining semantic consistency for corresponding tokens, while $\mathcal{L}_{ssim}$ preserves structural consistency. Their combination jointly enforces semantic and spatial constraints, leading to the best overall performance.

\begin{figure}[t]
\centering

\includegraphics[width=\linewidth]{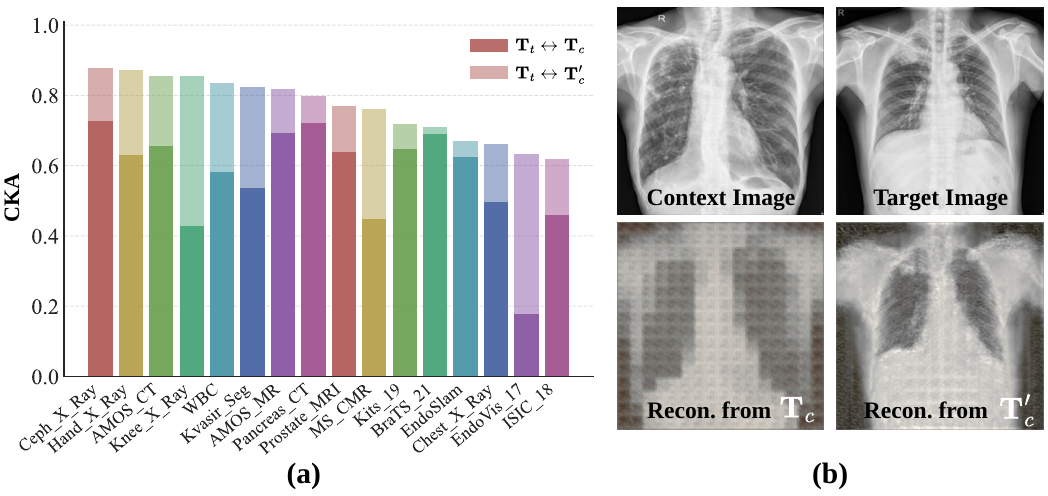}
\vspace{-5ex}
\caption{
(a) Feature structural similarity measured by CKA, where $\leftrightarrow$ indicates the feature pairs used for computation.
(b) Reconstruction results of the target image using $\mathbf{T}_c$ and $\mathbf{T}_c'$.
}
\label{fig:registration}
\end{figure}

\noindent \textbf{Robustness under large context-target variation.} Unlike common explicit pixel-level registration, our SRT performs implicit registration by computing token-level attention correspondences, enabling substantial spatial and semantic deformation. This makes the registration highly adaptive, rather than limited to cases where context and target images are highly similar. We visualize the features before and after registration under large variations (\eg, anatomical differences and tool placement changes) using PCA \cite{PCA} in Figure~\ref{fig:large_variation}. SRT generalizes well across such variations and precisely aligns context tokens to the target ones.

\noindent \textbf{Assessing Registration.}
We assess the effectiveness of registration by measuring the structural similarity of features pre- and post-registration, and by examining the quality of the reconstructed images.
We first compute the Centered Kernel Alignment (CKA) \cite{CKA} between $\mathbf{T}_t$ and $\mathbf{T}_c$, $\mathbf{T}_c'$, respectively, to measure their structural similarity in the feature space.
As shown in Figure~\ref{fig:registration}(a), the registered features $\mathbf{T}_c'$ exhibit consistently higher CKA values than $\mathbf{T}_c$ across almost all datasets, indicating a closer alignment with $\mathbf{T}_t$.

We further use the frozen model trained on MedGen to train a reconstructor that maps $\mathbf{T}_c$ and $\mathbf{T}_c'$ to the target image.
As illustrated in Figure~\ref{fig:registration}(b), the image reconstructed from $\mathbf{T}_c'$ is visually and spatially highly consistent with the target, while $\mathbf{T}_c$ remains similar to the context, thus fails to reconstruct meaningful structures.
This clearly validates that $\mathbf{T}_c'$ becomes highly aligned with the target after registration.
See Appendix \ref{sec:Assessing_Registration} for more visualizations.

\section{Conclusion}
In this paper, we introduced a novel task, termed General-purpose Medical Dense Prediction (GMDP). The idea of GMDP promotes the development of generalist medical vision systems that can assist a wide range of clinical activities, with potential applications such as AI-based surgical assistance.
We proposed a GMDP framework, Generalist Registration-Aware In-context Learner (GRAIL), and constructed a large-scale GMDP dataset, Medical Generalist (MedGen), which covers five mainstream dense prediction tasks across diverse medical scenarios.
Experimental results demonstrate that GRAIL outperforms baselines by a significant margin.

\noindent \textbf{Limitations and Future Work.}
This work focuses on pixel-level understanding, yet image-level task classification is not specifically considered.
While our model performs well on single-object Seg and OD, we remain optimistic about its potential for multi-object and instance-level perception. Moreover, multimodal prompting with MLLMs may further improve GMDP.
We hope this work will inspire further research toward broader task coverage and more unified visual modeling in the medical domain.

{
    \small
    \bibliographystyle{ieeenat_fullname}
    \bibliography{gmdp}
}

\FloatBarrier
\clearpage

\begin{figure*}[t]
    \centering
    \includegraphics[width=0.99\linewidth]{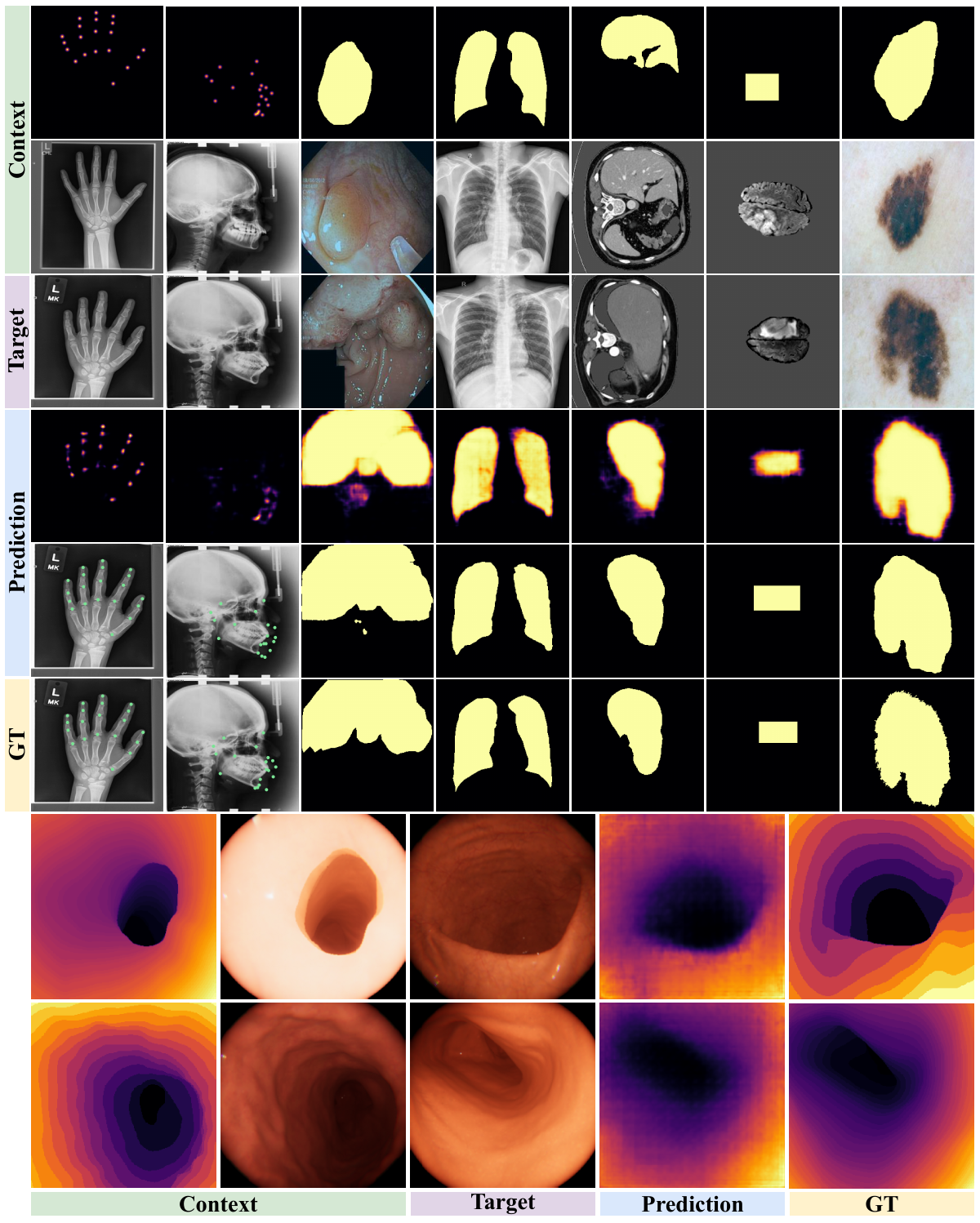}
    \caption{Additional In-context Inference Results by GRAIL}
    \label{fig:APPENDIX_VISUAL}
\end{figure*}

\FloatBarrier
\clearpage

\appendix
\beginappendixcites
  \startcontents[appendix]

  \section*{Appendix}

  \begingroup
 \dottedcontents{section}[2.5em]{\bfseries}{2.5em}{0.6em}
  \dottedcontents{subsection}[5.7em]{}{3.2em}{0.6em}

  \printcontents[appendix]{}{1}{\setcounter{tocdepth}{2}}
  \endgroup

\vspace{5ex}
In the following sections, we provide additional technical and experimental details, further analysis, and dataset descriptions. In addition, we present more qualitative results of GRAIL in Figure \ref{fig:APPENDIX_VISUAL}.

\section{More Diagnostic Experiments}

\begin{table}[t]
\centering
\caption{Task classification accuracy of intermediate representations. The classifier predicts the task type among DE, Seg, OD, LD, and OC.}
\label{tab:representation_accuracy}
\resizebox{0.85\linewidth}{!}{
\begin{tabular}{lc}
\toprule
\textbf{Representation} & \textbf{Accuracy (\%)} \\
\midrule
Target image \(\mathbf T_t\) & 53.21 \\
Context image \(\mathbf T_c\) & 52.56 \\
Context label \(\mathbf L_c\) & 70.83 \\
Context image + label & 79.12 \\
Registered \(\mathbf Z'\) after SRT & 49.96 \\
\bottomrule
\end{tabular}
}
\end{table}
\subsection{Does GRAIL Learn Task-agnostic Representations?}
To verify whether GRAIL learns a unified input-output mapping rather than simply memorizing task-specific data patterns, we perform linear probing to examine whether its intermediate features contain task information. We train linear classifiers on frozen representations from different stages of GRAIL, where lower task classification accuracy indicates more task-agnostic and generalizable representations, containing less task-related information.

As shown in Table~\ref{tab:representation_accuracy}, individual image representations contain limited task information, while the context label provides stronger task cues since it directly specifies the prediction target. Combining context image and label features further improves task prediction accuracy to 79.12\%, demonstrating that GRAIL effectively uses the context pair as a visual prompt to implicitly identify the task.

After SRT, task predictability decreases to 49.96\%, indicating that SRT transforms the task-aware context representation into a more task-agnostic and spatially aligned feature space. In other words, SRT successfully transforms context representations into a more task-agnostic and spatially aligned feature space, which enables the following decoder to perform unified prediction across diverse tasks. This supports our motivation that GRAIL achieves context-conditioned inference through spatial registration rather than relying on fixed task-specific semantics.

\begin{figure*}[t]
    \centering
    \includegraphics[width=0.99\linewidth]{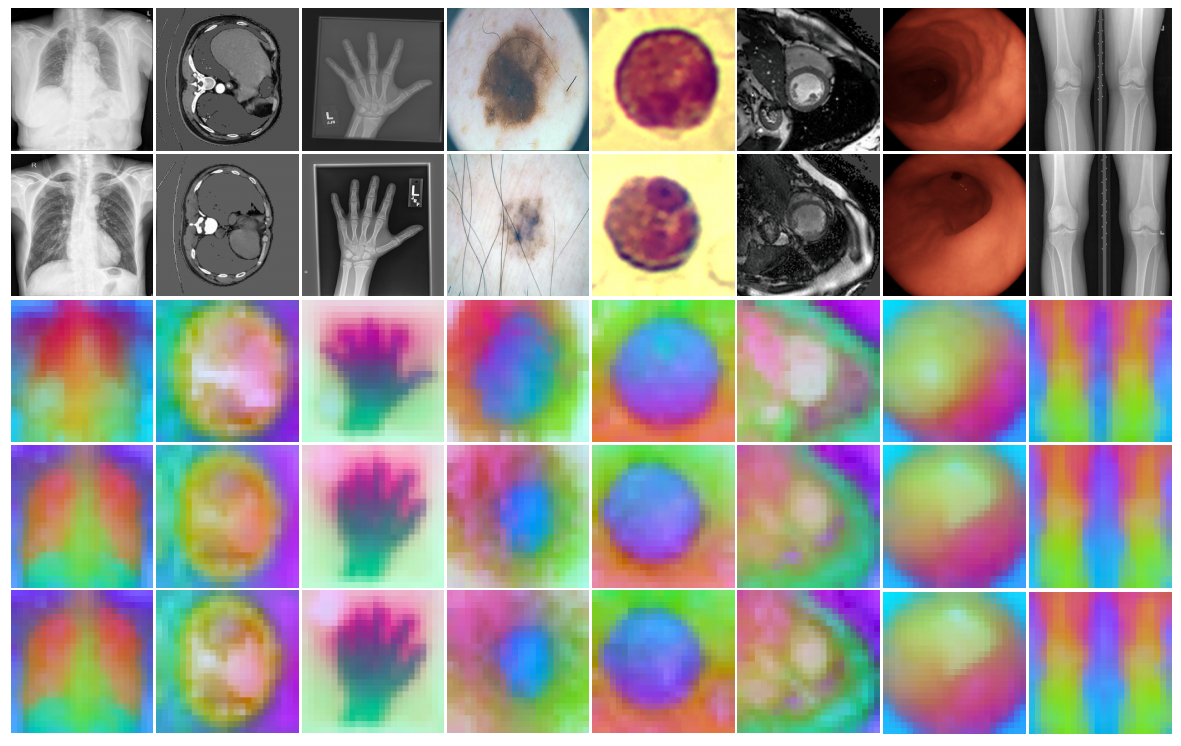}
    \caption{Visualization of feature registration results using PCA \cite{PCA}.
From top to bottom, each row shows the context image, target image, unregistered tokens $\mathbf{T}_{c}$, registered tokens $\mathbf{T}_{c}'$, and target tokens $\mathbf{T}_{t}$. Regions with similar semantics across different images exhibit similar colors. We can conclude that the registered features not only share structural similarity with the target but also achieve semantic alignment.}
    \label{fig:PCA_reg}
\end{figure*}

\subsection{Context Selection / Robustness to Different Levels of Context–Target Discrepancy}
In Section~\ref{sec:ablations}, we qualitatively evaluate the robustness of our model under large context–target discrepancy. In this section, we further provide a quantitative analysis of the model's robustness across different levels of context–target variation. We adopt SSIM \cite{SSIM} as an image similarity metric and, during inference, search for context images with different similarity levels to each target image. Note that this setting is used solely for analysis purposes; we do not recommend it for practical deployment due to the additional search overhead. According to the SSIM scores, we define two thresholds (SSIM = 0.50 and 0.80) and construct three levels of context–target discrepancy: large discrepancy (SSIM $<$ 0.50), moderate discrepancy (0.50 $\le$ SSIM $\le$ 0.80), and small discrepancy (SSIM $>$ 0.80).

\begin{table}[t]
\centering
\caption{Performance across context-target discrepancy levels.}
\label{tab:discripency_level}
\resizebox{0.9\linewidth}{!}{
\begin{tabular}{lcccc}
\toprule
\multicolumn{1}{c}{\multirow{2}{*}{\begin{tabular}[c]{@{}c@{}}\textbf{Discrepancy}\\ \textbf{Level}\end{tabular}}} & \multicolumn{4}{c}{\textbf{Task}}      \\ \cmidrule(lr){2-5}
\multicolumn{1}{c}{}                                                                             & DE$\downarrow$    & Seg$\uparrow$    & OD$\uparrow$    & LD$\uparrow$    \\ \midrule
Random & 0.019 & 72.40 & 46.18 & 82.19\\
\midrule
Large                                                                                            & 0.027 & 70.12 & 44.82 & 81.31 \\
Moderate                                                                                         & 0.021 & 72.81 & 46.14 & 81.69 \\
Small                                                                                            & \textbf{0.015} & \textbf{73.85} & \textbf{47.95 }& \textbf{82.54} \\ \bottomrule
\end{tabular}
}
\end{table}

As shown in Table~\ref{tab:discripency_level}, our model maintains strong performance across different levels of context–target discrepancy, indicating high robustness to variation. Note that `Random' denotes the context selection strategy used in our main experiments. While larger discrepancies inevitably introduce a slight performance drop, particularly for tasks with substantial scene variation such as DE, the degradation remains minimal. This demonstrates that the proposed method remains stable even under challenging data conditions, without requiring high-quality context examples. On the other hand, the optimal performance observed under small variations indicates that, with an appropriate context selection strategy to identify more informative context examples from the context pool, the in-context inference performance of GRAIL can be further improved.

\subsection{More on Assessing Registration.}
\label{sec:Assessing_Registration}
In Section \ref{sec:ablations}, we assess the effect of registration by reconstructing the target image.
We provide the implementation details of the reconstruction process and directly visualize the feature representations before and after registration in this section.
We adopt an 8-layer ViT reconstructor, to reconstruct the target image from the context tokens.
Two separate training runs are performed: one using the unregistered features $\mathbf{T}_c$, and the other using the registered features $\mathbf{T}_c'$.
The reconstructor is trained on the Chest X-Ray~\cite{Chestxray1,Chestxray2} training set with a combination of L1 and SSIM losses, and evaluated on its validation set.

We directly visualize the image tokens in Figure~\ref{fig:PCA_reg} by mapping them into the RGB space using PCA~\cite{PCA}. As shown, the registered context tokens $\mathbf{T}_c'$ are spatially well aligned with the target tokens $\mathbf{T}_t$, and the corresponding regions exhibit nearly identical colors, indicating strong semantic consistency.
Unlike traditional registration methods based on explicit deformation fields~\cite{MIA_DEFORM_REG,VOXELMORPH}, our approach performs attention-based implicit semantic deformation, which allows more flexible alignment, especially when the target and context images exhibit large structural or appearance variations that conventional methods often fail to handle.

\begin{figure}[t]
    \centering
    \includegraphics[width=0.99\linewidth]{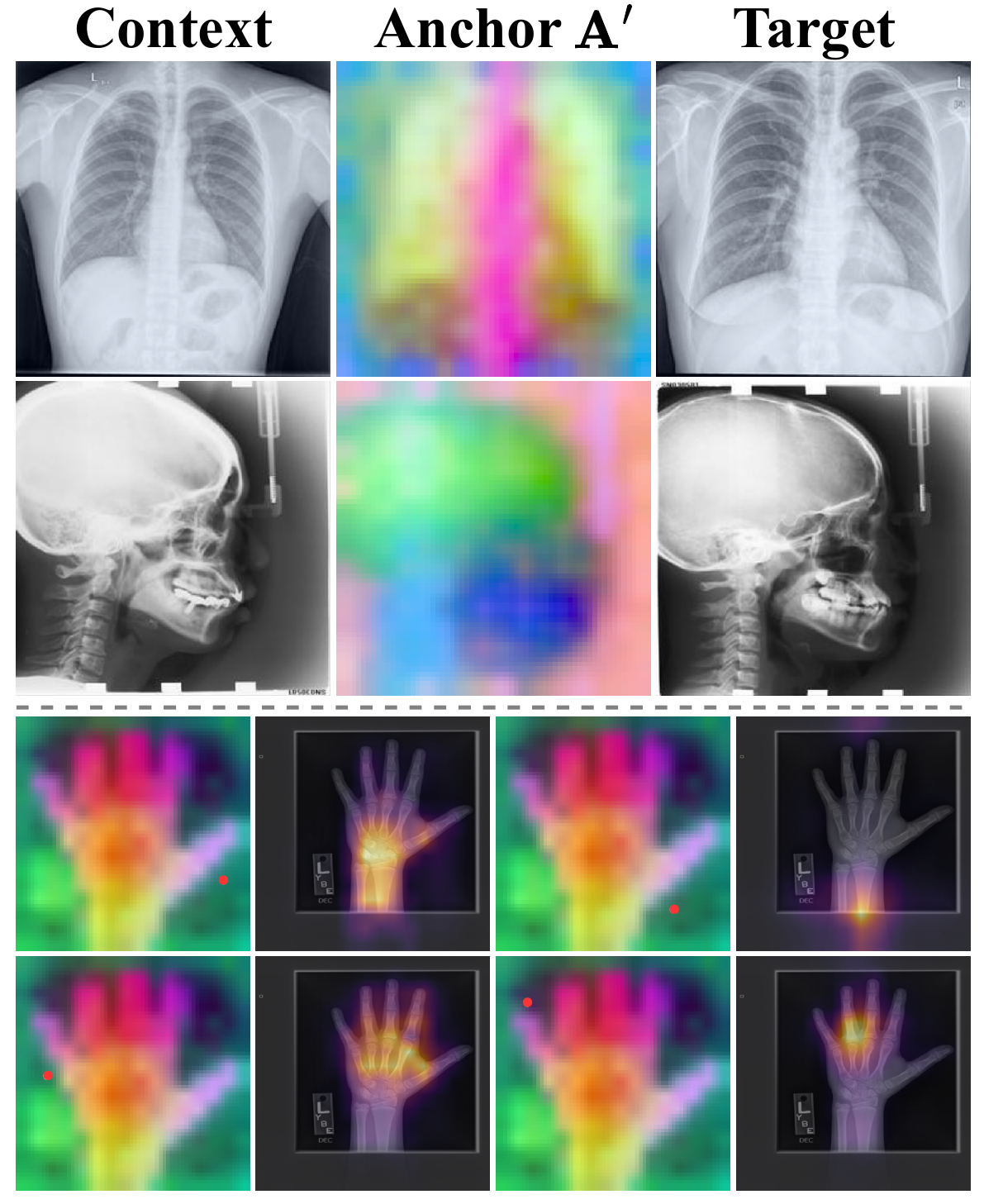}
    \caption{Visual interpretation of the deformation anchor $\mathbf{A}'$. Top part: visualization of $\mathbf{A}'$. Bottom part: visualization of the attention maps generated from selected anchor tokens.}
    \label{fig:deformation_anchor}
\end{figure}

\subsection{Visual Interpretability of Deformation Anchor}
In Section \ref{sec:ASP}, we instance-adaptively generate a deformation anchor $\mathbf{A}'$ from a set of directional vectors $\mathbf{V}$ that indicate the context–target deformation displacement. The anchor $\mathbf{A}'$ is then used to guide the deformation of the context toward the target. In this section, we visualize two components: the anchor $\mathbf{A}'$ itself and the interaction between the tokens in $\mathbf{A}'$ and the context tokens, as shown in Figure \ref{fig:deformation_anchor}.

Figure \ref{fig:deformation_anchor} top part visualizes $\mathbf{A}'$ using PCA \cite{PCA}. It exhibits spatial patterns from both the context and the target, with a stronger tendency toward the target. This indicates that it perceives the context–target displacement encoded in $\mathbf{V}$ and contains sufficient target information to guide the deformation towards the target.

Figure \ref{fig:deformation_anchor} bottom part visualizes the interactions between selected tokens in $\mathbf{A}'$ and the context tokens $\mathbf{Z}$. The cross-attention heatmaps show that different anchor tokens attend to distinct spatial regions, where some focus on local fine-grained details and others integrate broader coarse-grained structures. This aligns with how humans register two similar images by considering both global alignment and local refinement. Through these interactions, $\mathbf{Z}$ absorbs the deformation cues and shifts toward the target, which we elaborate on in Section \ref{sec:Assessing_Registration}.

\subsection{Effect of Encoder Choice}

We investigate different encoder choices for the image encoder $f_{I}(\cdot)$ and the label encoder $f_{L}(\cdot)$.
For $f_{I}(\cdot)$, we adopt DINOv3 \cite{DINOV3} backbones of different sizes, including ViT-S, ViT-B, and ViT-L.
As shown in Table~\ref{tab:backbone_choice}, ViT-B achieves the overall best performance among all variants. We attribute this to the fact that the ViT-B architecture preserves finer local structures while maintaining sufficient representation capacity.
Overall, all three variants perform consistently well, demonstrating the encoder-agnostic robustness of our framework.
For $f_{L}(\cdot)$, we examine ViT-B and ViT-L with two initialization strategies: ImageNet \cite{imagenet} pretraining and training from scratch.
The results show that $f_{L}(\cdot)$ remains stable across tasks except for DE and LD, where fine-grained label understanding is required and thus the encoder exerts a stronger influence.
Overall, model size has only a limited impact on the final results, while the ImageNet-pretrained variant performs slightly better, despite being used for label encoding rather than natural image recognition.

\begin{table}[t]
\centering
\caption{Effect of Encoder Choice across DINOv3 Variants and ViT Initializations.}
\label{tab:backbone_choice}
\resizebox{0.99\linewidth}{!}{
\begin{tabular}{c|c|cccc}
\toprule
Encoder & Version & DE$\downarrow$ & Seg$\uparrow$ & OD$\uparrow$ & LD$\uparrow$ \\
\midrule
\multirow{3}{*}{$f_{I}(\cdot)$}
    & ViT-L & 0.021 & \textbf{72.95} & 45.93 & 79.31 \\
    & ViT-B & \textbf{0.019} & 72.39 & \textbf{46.18} & \textbf{82.19} \\
    & ViT-S & 0.032 & 70.81 & 42.47 & 77.72 \\
\midrule
\multirow{4}{*}{$f_{L}(\cdot)$}
    & ViT-B-From scratch & 0.034 & 72.13 & 45.00 & 76.85 \\
    & ViT-B-ImageNet & \textbf{0.019} & \textbf{72.39} & 46.18 & 82.19 \\
    & ViT-L-From scratch & 0.025 & 71.23 & 45.92 & 79.77 \\
    & ViT-L-ImageNet & 0.021 & 72.22 & \textbf{46.87} & \textbf{83.14} \\

\bottomrule
\end{tabular}
}
\end{table}

\begin{figure}[t]
\centering
\includegraphics[width=0.99\linewidth]{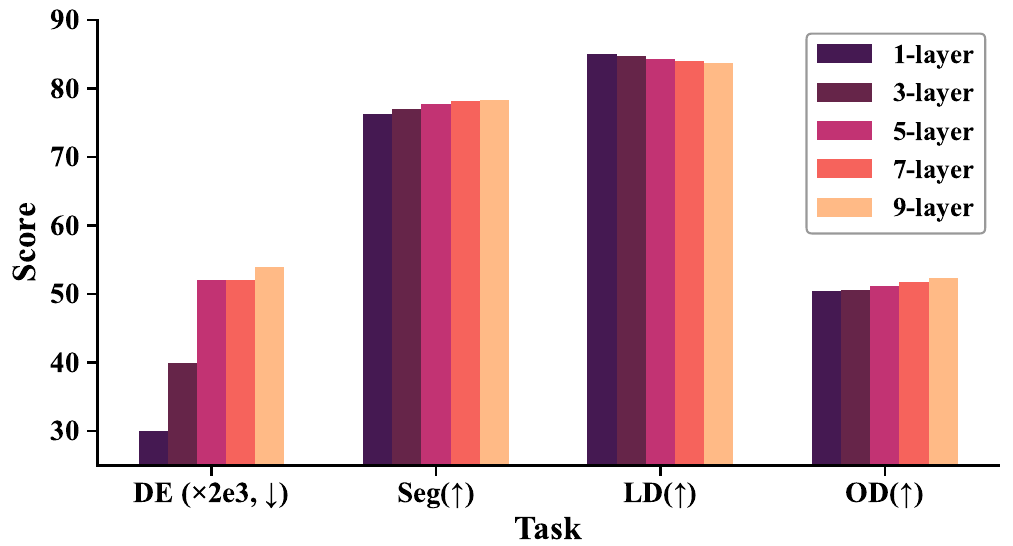}
\caption{Results of different numbers of SRT layers on various tasks.}
\label{fig:SRT_layers}
\end{figure}

\subsection{SRT used as Stacked Layers}
SRT can be viewed as a unified layer that can be directly stacked at multiple depths of the network to iteratively process intermediate features.
This enables progressive spatial perception and semantic deformation, which may lead to more precise registration results.
We implement this idea and report the results in Figure~\ref{fig:SRT_layers}.
We observe that the overall performance is relatively insensitive to the number of SRT layers.
Specifically, the results on Seg and OD gradually improve with increasing layer depth and then saturate.
We attribute this to the fact that these tasks typically exhibit larger context–target variations, where multiple deformations enable more accurate and sufficient registration.
This also indicates that SRT can serve as a powerful plug-and-play module for reducing inter-sample variations in such tasks.
In contrast, DE and LD show a slight decline as the number of layers increases, since these tasks rely on highly discriminative token representations, while deeper attention aggregation tends to smooth inter-token semantics.

\subsection{Efficiency Analysis.}
From Table \ref{tab:efficiency}, we conclude that our method also maintains high efficiency. It has the minimal training time cost and requires only slightly more memory and inference time than the previous baseline UniAP \cite{UNIAP}, while being significantly more efficient than MAE-based methods \cite{PAINTER,MAEINPAINTING}. This makes it more suitable for deployment on medical devices.

\begin{table}[t]
\centering
\setlength{\tabcolsep}{6pt}
\renewcommand{\arraystretch}{1.15}
\caption{Efficiency comparison of training and inference cost.}
\label{tab:efficiency}

\resizebox{\linewidth}{!}{
\begin{tabular}{lcccc}
\toprule
\multirow{3}{*}{Method} & \multicolumn{2}{c}{Train} & \multicolumn{2}{c}{Inference} \\
\cmidrule(lr){2-3} \cmidrule(lr){4-5}
& Latency & Memory & Latency & Memory \\
& \scriptsize (ms/img) & \scriptsize (GB) & \scriptsize (ms/img) & \scriptsize (GB) \\
\midrule
VTM \cite{VTM} & \underline{68.44} & 6.65 & 15.57 & 3.62 \\
UniAP \cite{UNIAP} & 84.44 & \textbf{2.59} & \textbf{9.43} & \textbf{1.79} \\
MAE-VQGAN \cite{MAEINPAINTING} & 78.22 & 3.91 & 23.52 & 1.98 \\
Painter \cite{PAINTER} & 129.17 & 9.46 & 36.83 & 4.05 \\
prompt-SeIF \cite{LZC_VICL} & 78.22 & 3.91 & 23.52 & 1.98 \\
SegMIC \cite{SEGMIC} & 49.21 & \underline{2.93} & 11.45 & 1.89 \\
 \rowcolor{blue!8}
\textbf{GRAIL (Ours)} & \textbf{47.89} & 3.01 & \underline{10.55} & \underline{1.81} \\
\bottomrule
\end{tabular}
}
\begin{tablenotes}
\item ``Memory'' reports the peak memory usage. All experiments use the same 4080 SUPER GPU. \end{tablenotes}
\end{table}

\begin{table}[h]
\centering
\caption{Effect of different numbers of context examples.}
\label{tab:multi_context}
\resizebox{0.9\linewidth}{!}{
\begin{tabular}{lcccc}
\toprule
\multicolumn{1}{c}{\multirow{2}{*}{\begin{tabular}[c]{@{}c@{}}\textbf{Context}\\ \textbf{Setting}\end{tabular}}}
& \multicolumn{4}{c}{\textbf{Task}} \\ \cmidrule(lr){2-5}
\multicolumn{1}{c}{}
& DE$\downarrow$ & Seg$\uparrow$ & OD$\uparrow$ & LD$\uparrow$ \\
\midrule
$K=1$
& 0.019 & 72.40 & 46.18 & 82.19 \\
$K=5$ Output
& 0.019 & 73.98 & 47.37 & 82.47 \\
$K=5$ Context
& 0.022 & 65.18 & 46.76 & 81.92 \\
$K=10$ Output
& \textbf{0.018} & \textbf{74.82} & \textbf{47.87} & \textbf{82.59} \\
$K=10$ Context
& 0.022 & 62.44 & 46.45 & 81.91 \\
\bottomrule
\end{tabular}
}
\end{table}

\subsection{Scaling In-context Inference with Additional Visual Prompts}
We evaluate the effect of increasing the number of context pairs with \(K=5,10\) to investigate whether additional visual prompts can further improve performance. We directly apply the model trained with \(K=1\) for in-context inference with more examples, without any retraining. To identify an effective strategy for scaling context examples, we consider two aggregation approaches. The first is \textbf{output-level} aggregation, where each context pair independently performs a forward pass with the target image, and the final prediction is obtained by averaging all predictions. The second is \textbf{context-level} aggregation, where the features of \(K\) context pairs are averaged before being used for inference.

As shown in Table~\ref{tab:multi_context}, output-level aggregation provides a more effective way to scale visual prompts, while the model consistently maintains or improves performance across all tasks with additional context examples. This demonstrates that GRAIL can further improve its in-context inference capability when provided with richer visual prompts.

\subsection{Failure Cases}
\label{sec:failure_cases}
\begin{figure}[t]
    \centering
    \includegraphics[width=\linewidth]{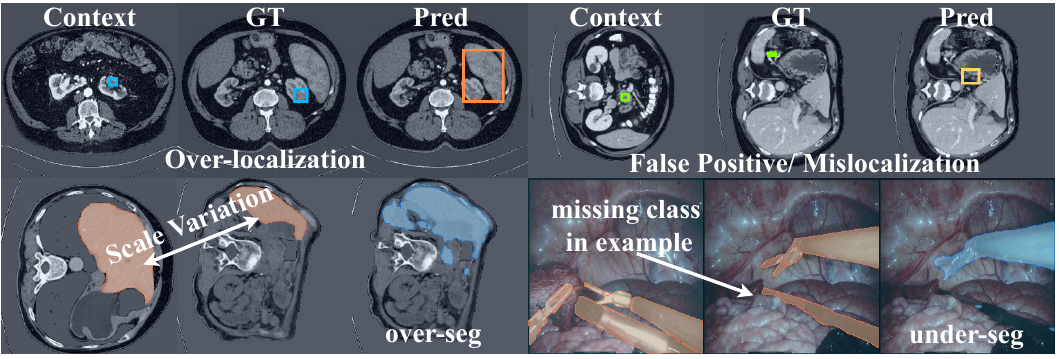}
    \caption{Failure cases on tiny targets (upper) and context-target mismatch (lower).}
    \label{fig:failure_case}
\end{figure}

Although our model achieves consistently strong performance on most targets, we observe two major failure modes: extremely small targets (\eg, tumors) and substantial discrepancies between the context and target images, as shown in Figure~\ref{fig:failure_case}. We primarily attribute these failures to an inherent bottleneck in VICL’s feature-matching nature. When the target is excessively small, context information may not be transferred accurately; when the context–target discrepancy is too large, the context example provides less effective guidance. Future work could incorporate attention mechanisms or loss functions specifically designed for small targets. In addition, improved context-example selection strategies could retrieve more relevant and informative samples from the context pool, thereby mitigating failures caused by large context–target discrepancies.

\section{Task Descriptions and Additional Implementation Details}
\label{sec:appendix_Task_Descriptions}
Our model produces \textit{unified} outputs across different tasks by generating probability maps or continuous value maps that follow the distribution of the context labels. We provide additional implementation details regarding the model’s output post-processing.
Note that post-processing is only performed for those tasks that involve discretization (\eg, segmentation or detection). All post-processing operations are easy to implement, avoiding the complex task-specific processing pipelines that are typically involved in conventional paradigms.

\noindent \textbf{Foreground Segmentation.}
Foreground Seg aims to distinguish the target region from the background and is one of the most fundamental tasks in medical image analysis. Its value primarily lies in boundary precision and quantitative interpretability, enabling direct applications in diagnosis, treatment planning, and disease monitoring \cite{MEDSAM}.
The binary foreground mask $\tilde{\mathbf{M}}$ is obtained by thresholding the output $\tilde{\mathbf{y}_t}$ at 0.5, \ie,
$\tilde{\mathbf{M}} = \mathbb{I}[\tilde{\mathbf{y}_t} \ge 0.5]$.

\noindent \textbf{Object Detection.}
In real clinical workflows, clinicians first need to identify roughly where abnormalities are located in an image and whether they require priority attention, \ie, retrieval first, followed by subsequent quantification \cite{nnDetection}. Medical OD therefore serves as an essential step in the diagnostic pipeline, substantially reducing the search space by enabling triage and candidate generation. Pixel-level predictions can then be performed on these key candidates to obtain precise quantitative measurements for treatment planning and longitudinal monitoring. This process also reduces the uncertainty and false positives/negatives that may arise when directly producing pixel-level masks for Seg, LD or related tasks \cite{LUNA16}.
In the GMDP setting, we consider the single-object detection scenario, consistent with prior visual in-context learning methods \cite{MAEINPAINTING,PAINTER,SD_VICL}. Following common practice in medical OD \cite{nnDetection,MOD_USING_SEG,MOD_USING_SEG_2}, we additionally annotate OD bounding boxes based on pixel-level label maps from segmentation benchmarks, as such datasets constitute the most widely available resources in medical imaging \cite{LENA_SEG_IS_MOST_STUDIED,nnDetection}.
Since OD training is identical to segmentation except that the context label is a binary box mask, we binarize the output map at 0.5 and extract ($\mathrm{BBox}(\cdot)$) the minimal bounding box $\tilde{\mathbf{B}}$ of the foreground area, \ie,
$\tilde{\mathbf{B}} = \mathrm{BBox}(\mathbb{I}[\tilde{\mathbf{y}_t} \ge 0.5])$.

\noindent \textbf{Landmark Detection.}
LD aims to localize keypoints in medical images, supporting tasks such as registration, precise measurement, and cross-frame surgical video object tracking. In particular, in the era of foundation models, accurate landmark annotations can directly serve as prompts indicating important structures or reference positions, enabling image reasoning and structured analysis. In our setting, the context label is represented as a Gaussian heatmap \cite{HRNET}, where each Gaussian kernel is centered at a landmark location.
We extract the coordinates of the top-$k$ local maxima from the output, where $k$ is determined by the number of context landmarks to ensure flexibility across different scenarios. Details are illustrated in Algorithm \ref{alg:keypoint_extraction}.

\begin{algorithm}[h]
\caption{Keypoint Extraction from Output}
\label{alg:keypoint_extraction}
\small
\begin{algorithmic}[1]
\Require Predicted output $\tilde{\mathbf{y}}_t$, target number $k$
\Ensure Keypoint coordinates $\mathbf{P}=\{p1,p2,\dots,p_k\}$

\State Initialize threshold $\tau=0.5$
\Repeat
    \State Find local maxima of $\tilde{\mathbf{y}}_t$ within a window size of $3$
    \State Keep points with $\tilde{\mathbf{y}}_t(x,y) \ge \tau$
    \State Decrease $\tau$ if fewer than $k$ points remain
\Until{$|\mathbf{P}| \ge k$}
\If{$|\mathbf{P}| > k$}
    \State Keep top-$k$ coordinates with the highest responses in $\tilde{\mathbf{y}}_t$
\EndIf
\State \Return $\mathbf{P}$
\end{algorithmic}
\end{algorithm}

\noindent \textbf{Depth Estimation.}
DE bridges 2D images to 3D geometric perception, which is particularly important in endoscopic and minimally invasive surgical scenarios, enabling accurate navigation and scene interaction for surgical tools and robotic systems. No post-processing is needed for this task.

\noindent \textbf{Object Counting.}
OC aims to estimate the number of target instances in images, such as cells. Inspired by conventional density-based counting methods that estimate object counts by integrating the density map \cite{sum_density_map_for_count}, we obtain the predicted count $\tilde{c}$ by integrating the output density map $\tilde{\mathbf{y}_t}$ and normalizing it with the context label statistics, \ie,
$\tilde{c} = \frac{\sum \tilde{\mathbf{y}_t}}{\sum \mathbf{y}_c / c_c},$
where $c_c$ denotes the context object count.

\vspace{1ex}
\noindent \textbf{Task Heterogeneity in MedGen.}
In Figure \ref{fig:MedGen} (b) we show that each task in MedGen has a different focus. From the perspective of data characteristics, OD emphasizes the localization of highly irregular or small structures such as tumors, whereas Seg focuses on boundary delineation across diverse categories and modalities. LD predicts sparsely distributed discrete points, while DE requires densely predicting the depth value for every pixel. We further include OC as a focus type unseen during training, which requires predicting the corresponding density map of an image. Each task plays an irreplaceable role in real biomedical imaging workflows.

A strong GMDP method should unify the processing of different tasks despite the heterogeneity among them. Our GRAIL adopts a general registration operation and implicitly encodes labels as features that interact contextually with image features, avoiding the introduction of task-specific inductive bias and thus unifying all tasks in context.

\begin{table*}[t]
\centering
\caption{Overview of datasets used in MedGen, including task type, imaging modality, region of interest (ROI), annotated categories, and total sample count.}
\label{tab:dataset_detail}
\rowcolors{2}{gray!10}{white}
\resizebox{\linewidth}{!}{
\begin{tabular}{lll>{\raggedright\arraybackslash}p{0.18\linewidth}>{\raggedright\arraybackslash}p{0.18\linewidth}r}
\toprule
\textbf{Dataset} & \textbf{Task} & \textbf{Modality} & \textbf{ROI} & \textbf{Categories} & \textbf{Total Images} \\
\midrule
Chest X-Ray \cite{Chestxray1,Chestxray2}      & Seg/OD & X-ray        & Chest                     & Lungs   & 567   \\
Hand X-Ray \cite{hand_knee_xr}      & LD     & X-ray        & Hand                      & Anatomical landmarks& 1833  \\
Knee X-Ray \cite{hand_knee_xr}      & LD     & X-ray        & Knee                      & Anatomical landmarks& 1015  \\
Cephalometric X-Ray \cite{ceph_xr} & LD  & X-ray        & Head                      & Anatomical landmarks& 400   \\
EndoVis17 \cite{ENDOVIS17}       & Seg/OD & Endoscopy    &        & Surgical instruments     & 525   \\
ISIC18 \cite{isic1,isic2}         & Seg    & Dermoscopy   & Skin lesions                & Melanoma, nevus, Seborrheic keratosis& 2594  \\
Kvasir-Seg \cite{kvasirSEG}      & Seg/OD & Endoscopy    & Gastrointestinal Tract         & Polyp      & 1000  \\
BraTS21 \cite{BRATS21}      & Seg& MRI          & Brain & Tumor (whole)& 24754 \\
AMOS-CT \cite{AMOS}     & Seg/OD & CT           & Abdomen                    & Liver, spleen, right kidney, left kidney      & 15919 \\
WBC  \cite{WBC}            & Seg    & Microscopy   &   —         & White blood cells      & 600   \\
MS-CMR  \cite{cmr,CMR2019}   & Seg& MRI          & Heart                      & Left ventricle blood pool, left ventricle myocardium, right ventricle      & 430   \\
AMOS-MRI \cite{AMOS}     & Seg/OD & MRI          & Abdomen                    & Liver, spleen, right kidney, left kidney      & 5525  \\
Kits19 \cite{kits19_1,kits19_2} & OD& CT& Abdomen                    & kidney tumor& 2701\\
Pancreas-CT \cite{MSD} & OD& CT& Abdomen                    & Pancreatic parenchyma, tumor& 7978\\
Prostate-MRI \cite{MSD} & OD& MRI& Prostate& Prostate peripheral zone, transition zone& 469\\
EndoSLAM \cite{EndoSLAM}     & DE     & Endoscopy    & Colon, Stomach, Small Intestine & — & 10797 \\
CBC   \cite{CBC}           & OC     & Microscopy   & —                          & Red blood cells, white blood cells      & 140   \\
\bottomrule
\end{tabular}
}
\end{table*}

\section{MedGen Details}
\label{sec:medgen_detail}
\subsection{Dataset Coverage}
We provide the details of each dataset in MedGen in Table~\ref{tab:dataset_detail}, including the associated task, image modality, region of interest (ROI), annotated categories, and the total number of images.

\begin{discussionbox}
\noindent \textbf{Data Imbalance or Inherent Task Distribution?}
A core contribution of MedGen is its substantially broader task coverage compared with existing single-task datasets. To construct such a comprehensive benchmark, we collected datasets with dedicated annotations for their original tasks whenever possible. However, owing to the scarcity of dedicated OD datasets, we also followed the \textbf{common practice} established in prior studies \cite{nnDetection,MOD_USING_SEG,MOD_USING_SEG_2} and generated OD annotations from segmentation masks. This strategy enables MedGen to contain 44.9\% OD data, but also results in a relatively high proportion of samples originating from segmentation datasets. Such an imbalance is \textbf{imposed} by the \textbf{nature} of medical imaging, where segmentation accounts for more than 70\% of existing challenges, as discussed in Section \ref{sec:appendix_Task_Descriptions}.

Moreover, although segmentation-originated data constitute the majority of MedGen, we find that they improve performance on smaller datasets, even across different tasks. Artificially removing 50\% of the segmentation data solely for \textit{task balancing} consistently degrades performance. For example, on the Cephalometric X-Ray dataset, an LD dataset containing only 400 images, removing segmentation data reduces PCK@0.1 from 79.86 to 77.32. Therefore, discarding useful segmentation data merely to obtain a more balanced task distribution would be neither empirically beneficial nor methodologically justified.

\end{discussionbox}

\subsection{Complex Tumors}
\begin{figure}[!ht]
    \centering
    \includegraphics[width=\linewidth]{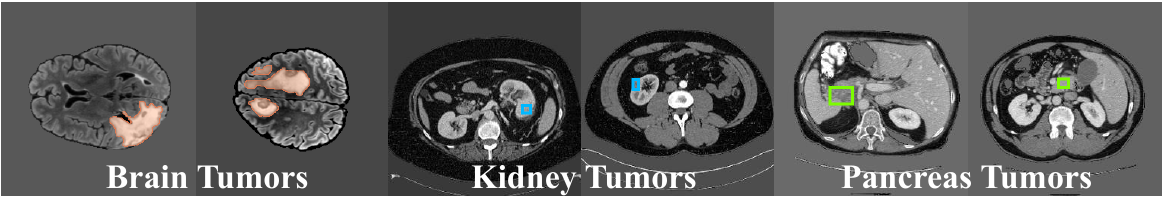}
    \caption{Examples of small and irregular tumors in MedGen, including brain, kidney, and pancreatic tumors. These cases exhibit substantial variations in size, shape, boundary, and image appearance.}
    \label{fig:complex_tumors}
\end{figure}
Beyond common structure-level segmentation/detection tasks involving abdominal or cardiac organs, MedGen includes several challenging tumor datasets that have conventionally been studied only by task-specific specialists (\cf Figure~\ref{fig:complex_tumors}). Although existing generalist models (\eg Painter \cite{PAINTER}) perform effectively across most common tasks, they underperform our model on complex tumor segmentation, as shown in Table~\ref{tab:SEG_DSC} and \ref{tab:OD} in the following section. Overall, our model achieves better performance on these challenging tasks. Nevertheless, we believe occasional failures on exceptionally complex tumor cases are reasonable, as discussed in {Section~\ref{sec:failure_cases}. For such corner cases, employing dedicated specialists rather than generalist models may be necessary to ensure trustworthy real-world deployment.}

\subsection{User-friendly Interface for MedGen}
\begin{figure}[t]
    \centering
    \includegraphics[width=0.99\linewidth]{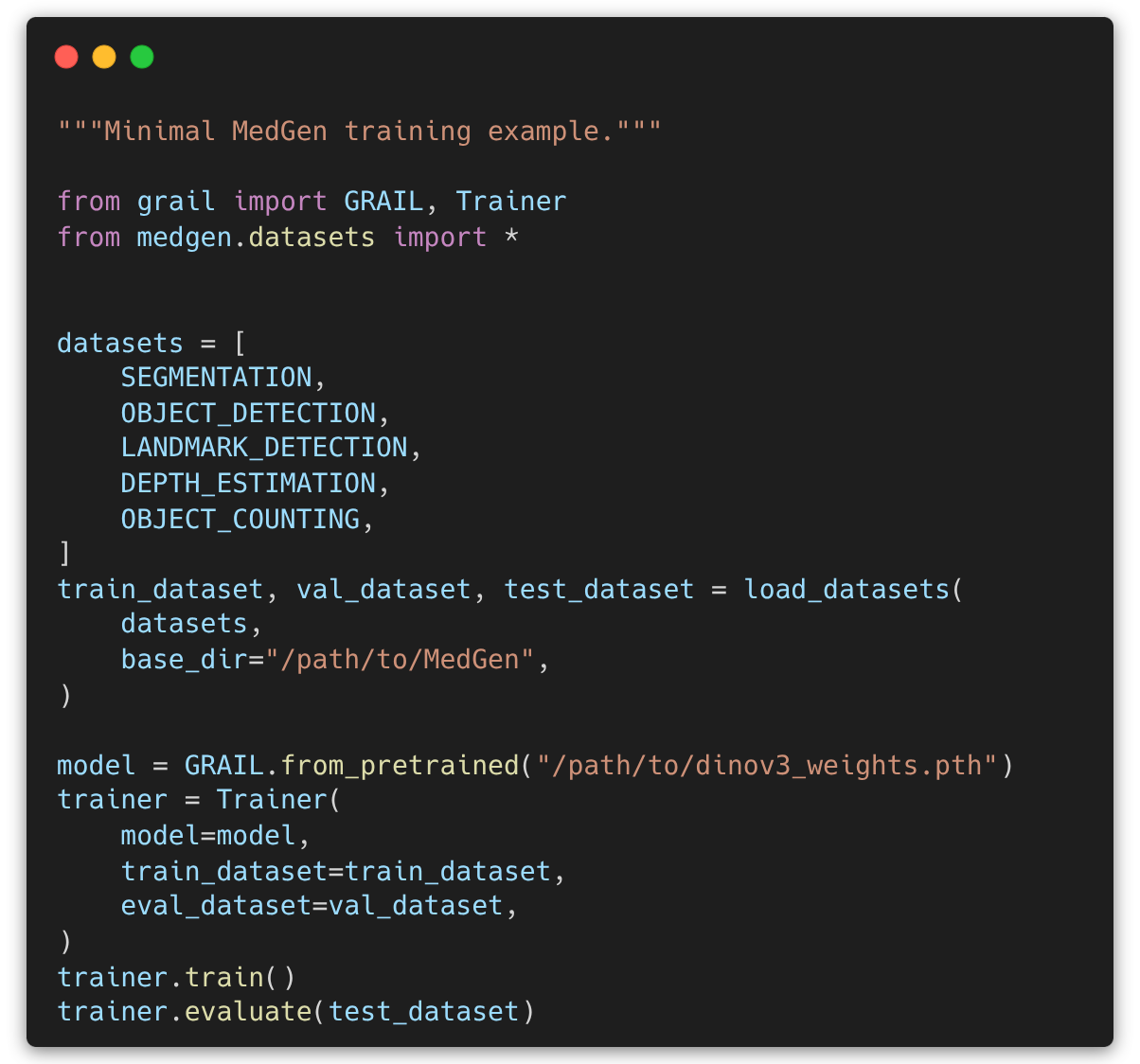}
    \caption{A minimal example of using the MedGen benchmark interface.}
    \label{fig:API}
\end{figure}

We make MedGen easy to access and build upon through a standardized data interface. Each source dataset is organized into the same structure, with a JSON file that describes task types, image paths, annotations, and categories. The data loader converts heterogeneous annotations into a common dense prediction format. We further encapsulate the underlying processing logic and provide a simple interface for accessing the benchmark, as illustrated in Figure \ref{fig:API}. Multiple datasets can therefore be used through the same workflow. We expect this unified organization and modular implementation to lower the barrier to using MedGen and accelerate future research on generalist medical vision models.

\begin{table*}[t]
\centering
\caption{Comparison of landmark detection results on individual datasets.}
\label{tab:LD}
\resizebox{\linewidth}{!}{
\begin{tabular}{l|cc|cc|cc|cc}
\toprule
\textbf{Method} &
\multicolumn{2}{c|}{\textbf{Hand X-Ray}} &
\multicolumn{2}{c|}{\textbf{Knee X-Ray}} &
\multicolumn{2}{c|}{\textbf{Cephalometric X-Ray}} &
\multicolumn{2}{c}{\textbf{Mean}} \\
\cmidrule(lr){2-9}& PCK@0.1$\uparrow$ & NMSE$\downarrow$ &
  PCK@0.1$\uparrow$ & NMSE$\downarrow$ &
  PCK@0.1$\uparrow$ & NMSE$\downarrow$ &
  PCK@0.1$\uparrow$ &
NMSE$\downarrow$
\\
\midrule
VTM \cite{VTM} & \textbf{98.08} & 0.001& 64.53& 0.0292& 77.14& 0.0407& 79.92& 0.0236
\\
UniAP \cite{UNIAP} & 88.08& 0.003& 53.24& 0.0382& 76.31& 0.0401& 72.54& 0.0271
\\
MAE-VQGAN \cite{MAEINPAINTING} & 77.24& 0.005& 61.28& 0.0301& 66.32& 0.0521& 68.28& 0.0291
\\
Painter \cite{PAINTER} & 81.51& 0.004& 74.62& \textbf{0.0212} & 56.52& 0.0459& 70.88& 0.0237
\\
prompt-SelF \cite{LZC_VICL}& 84.71& 0.004& 71.95& 0.0251& 62.35& 0.0537& 73.00&
0.0276\\
UniverSeg \cite{UNIVERSEG}& 75.48& 0.006& 61.81& 0.0441& 59.18& 0.0428& 65.49& 0.0310\\
Tyche \cite{Tyche}& 96.14& 0.001& \textbf{74.95}& 0.0224& 71.80& 0.0321& 80.96&

0.0185\\
SegMIC \cite{SEGMIC}& 79.51& 0.005& 72.32& 0.0249& 61.25& 0.0592& 71.03&

0.0297\\
 \rowcolor{blue!8}
\textbf{GRAIL (Ours)} & 97.63& \textbf{0.0007} & 69.07& 0.0235& \textbf{79.86} & \textbf{0.0170} & \textbf{82.19} &
\textbf{0.0137}
\\
\bottomrule
\end{tabular}
}
\end{table*}

\begin{table*}[t]
\centering
\caption{Comparison of segmentation results on individual datasets. Performance is reported in DSC$\uparrow$.}
\label{tab:SEG_DSC}
\resizebox{\linewidth}{!}{
\begin{tabular}{l|ccccccc|c}
\toprule
\textbf{Method} &
\textbf{AMOS-CT} & \textbf{BraTS21} & \textbf{MS-CMR} &
\textbf{EndoVis} & \textbf{ISIC18} &
\textbf{Kvasir-Seg} & \textbf{WBC} & \textbf{Mean}
\\
\midrule
VTM \cite{VTM} & 59.25& \textbf{69.27} & 75.61& 69.42& 81.18& 79.40& 60.77& 70.70\\
UniAP \cite{UNIAP} & 48.85& 62.06& 67.33& 71.64& 74.56& 70.62& 61.17& 65.18
\\
MAE-VQGAN \cite{MAEINPAINTING} & 57.33& 63.29& 59.93& 65.98& 79.88& 71.02& \textbf{65.00} & 66.06
\\
Painter \cite{PAINTER} & 52.34& 59.17& 66.82& 69.32& \textbf{83.74} & 78.33& 73.92& 69.09
\\
prompt-SelF \cite{LZC_VICL}& 60.81& 61.03& 70.18& 69.35& 79.66& 79.05& 64.58& 69.24\\
UniverSeg \cite{UNIVERSEG}& 38.89& 63.84& 48.94& 44.52& 61.21& 42.11& 56.74& 50.89\\
Tyche \cite{Tyche}& 49.19& 67.69& 61.83& 60.44& 66.69& 60.82& 62.56& 61.32\\
SegMIC \cite{SEGMIC}& 55.39& 63.28& 71.92& 65.33& 78.88& 68.71& 59.32& 66.12\\
 \rowcolor{blue!8}
\textbf{GRAIL (Ours)} & \textbf{62.89} & 68.01& \textbf{77.65} & \textbf{72.43} & 81.59& \textbf{81.20} & 63.01& \textbf{72.40}
\\
\bottomrule
\end{tabular}
}
\end{table*}

\section{Discussion on Baseline Coverage}
\label{sec:baseline_coverage}
In Section \ref{sec:Benchmark_result}, we benchmark against in-context learning methods designed for both natural \cite{VTM,UNIAP,MAEINPAINTING,PAINTER,LZC_VICL} and medical images \cite{Tyche,SEGMIC}, ensuring a broad coverage of model architectures and target scenarios. This enables a comprehensive evaluation and avoids potential bias from comparing only against multi-task methods lacking medical focus or medical methods tailored exclusively to segmentation. The results also highlight the advantage of GRAIL on GMDP by leveraging the structured nature of medical data in the pipeline design.
However, we acknowledge that several recent in-context learning works relevant to our study (also discussed in Section 2) are not included in our experiments. We clarify our rationale below.

\textbf{First,} for multi-modal in-context methods \cite{VICL4LVLM,MMICL}, MedGen does not currently support text grounding or other multi-modal inputs. Beyond the absence of rigorous or structured textual representations in our setting, we argue that visual prompts more naturally convey input–output mappings for dense prediction than text prompts, particularly for tasks such as DE, where pixel-wise outputs do not correspond to discrete semantic concepts \cite{SAM3}. This choice is also consistent with clinical practice, where densely annotated imaging data are typically more abundant \cite{nnDetection}. \textbf{Second,} we do not compare against several medical in-context \textit{segmentation} methods \cite{Show_and_Segment,ssl-alp2020,FAMNET,SINE} due to their strong task-specific inductive biases, \eg, deriving target-class representations by multiplying context labels with context features. Such formulations are tightly coupled with segmentation and fail to generalize to the broader GMDP setting (\eg, DE and OC). By contrast, GRAIL is designed with minimal task-specific design to move beyond specialized formulations toward more general medical intelligence. \textbf{Third,} several relevant works \cite{SD_VICL,Show_and_Segment,VICL_GLOBAL_OP} do not provide open-source implementations, which unfortunately prevents faithful reproduction and thus a fair evaluation on the proposed MedGen.

We therefore prioritize the most recent open-source baselines with minimal task-specific design, ensuring a comprehensive and well-justified evaluation.

\section{Results on Individual Datasets}
\label{sec:Results_on_Individual_Datasets}
In Table \ref{tab:main}, we only reported the average results across all datasets.
Here, we provide the detailed results on each individual dataset, as shown in Table \ref{tab:LD}, \ref{tab:SEG_DSC},  \ref{tab:HD95}, and \ref{tab:OD}.
We observe that different model architectures exhibit considerable performance variations across datasets.
For instance, our model performs particularly well on AMOS \cite{AMOS} for Seg, whereas the MAE-based method Painter achieves better results on ISIC18 \cite{isic1,isic2}.
Overall, however, our method consistently surpasses other approaches on the majority of datasets and shows outstanding performance especially on complex datasets containing multiple object categories within a single image, such as abdomen or cardiac datasets.

\begin{table*}[t]
\centering
\caption{Comparison of segmentation results on individual datasets. Performance is reported in HD95$\downarrow$.}
\label{tab:HD95}
\resizebox{\linewidth}{!}{
\begin{tabular}{l|ccccccc|c}
\toprule
\textbf{Method} &
\textbf{AMOS-CT} & \textbf{BraTS21} & \textbf{MS-CMR} &
\textbf{EndoVis} & \textbf{ISIC18} &
\textbf{Kvasir-Seg} & \textbf{WBC} & \textbf{Mean}
\\
\midrule
VTM \cite{VTM} & 33.23& 34.21& 16.11& 69.23& 28.07& 37.34& 37.15& 36.48
\\
UniAP \cite{UNIAP} & 67.24& 27.22& 26.21& \textbf{58.57} & 38.41& 49.00& 46.96& 44.80\\
MAE-VQGAN \cite{MAEINPAINTING} & 62.33& 48.92& 31.26& 72.91& 21.71& \textbf{24.53} & 24.25& 40.84
\\
Painter \cite{PAINTER} & 69.25& 54.81& 37.72& 64.26& \textbf{15.99} & 41.02& \textbf{21.33} & 43.48
\\
prompt-SelF \cite{LZC_VICL}& 47.55& 56.82& 35.82& 62.74& 39.25& 39.21& 38.94& 45.76\\
UniverSeg \cite{UNIVERSEG}& 70.19& 33.65& 41.3& 52.72& 52.14& 79.46& 31.94& 51.63\\
Tyche \cite{Tyche}& 69.42& 24.82& 37.79& 66.78& 57.74& 89.87& 43.15& 55.65\\
SegMIC \cite{SEGMIC}& 55.98& 49.60& 35.42& 72.91& 39.78& 77.00& 58.74& 55.63\\
 \rowcolor{blue!8}
\textbf{GRAIL (Ours)} & \textbf{32.36} & \textbf{26.61} & \textbf{15.59} & 64.34& 26.27& 36.10& 45.32& \textbf{35.23}
\\
\bottomrule
\end{tabular}
}
\end{table*}

\begin{table*}[t]
\centering
\caption{Comparison of object detection results on individual datasets. Performance is reported in mIoU$\uparrow$.}
\label{tab:OD}
\resizebox{\linewidth}{!}{
\begin{tabular}{l|ccccccc|c}
\toprule
\textbf{Method} &
\textbf{AMOS-CT} & \textbf{Chest X-Ray} & \textbf{EndoVis} &
\textbf{Kvasir-Seg} & \textbf{Kits19}&
\textbf{Pancreas-CT}& \textbf{Prostate-MRI}& \textbf{Mean}
\\
\midrule
VTM \cite{VTM} & 36.31& 48.66& 40.45& 53.08& 45.82& 29.55& \textbf{61.17}& 45.00\\
UniAP \cite{UNIAP} & 32.13& 53.56& 37.37& 52.28& 41.29& 25.71& 55.42& 42.54\\
MAE-VQGAN \cite{MAEINPAINTING} & 36.28& \textbf{56.31}& 39.33& 45.72& 45.91& 31.24& 58.46& 44.75
\\
Painter \cite{PAINTER} & 32.77& 51.12& 41.25& 48.99& 42.51& 24.17& 57.29& 42.59\\
prompt-SelF \cite{LZC_VICL}& 34.25& 55.87& 38.59& 53.24& 41.35& 31.85& 58.95& 44.87\\
UniverSeg \cite{UNIVERSEG}& 28.93& 24.81& 45.92& 47.47& 28.85& 33.6& 51.25& 37.26\\
Tyche \cite{Tyche}& 35.44& 52.88& 19.97& 52.55& 37.59& 29.95& 58.25& 40.95\\
SegMIC \cite{SEGMIC}& 31.22& 49.93& 39.88& 49.00& 42.36& 29.01& 55.70& 42.44\\
 \rowcolor{blue!8}
\textbf{GRAIL (Ours)} & \textbf{37.35}& 49.67& \textbf{41.62}& \textbf{54.12}& \textbf{47.29}& \textbf{33.19}& 60.04& \textbf{46.18}\\
\bottomrule
\end{tabular}
}
\end{table*}

\section{Evaluation Metrics}
\label{sec_evaluation_metric}
This section provides detailed definitions of the evaluation metrics used in our paper.

\noindent \textbf{Centered Kernel Alignment (CKA).}
CKA \cite{CKA} is a similarity metric that quantifies the alignment between two feature representations by comparing their centered Gram matrices, thereby capturing structural correspondence rather than direct value matching.
Formally, it is defined as the normalized Hilbert--Schmidt Independence Criterion (HSIC):
\begin{equation}
\text{CKA}(\mathbf{K},\mathbf{L})=
\frac{\operatorname{HSIC}(\mathbf{K},\mathbf{L})}
{\sqrt{\operatorname{HSIC}(\mathbf{K},\mathbf{K})\operatorname{HSIC}(\mathbf{L},\mathbf{L})}},
\end{equation}
where \(\mathbf{K} = \mathbf{H}\mathbf{X}\mathbf{X}^{\top}\mathbf{H}\) and \(\mathbf{L} = \mathbf{H}\mathbf{Y}\mathbf{Y}^{\top}\mathbf{H}\) are the centered Gram matrices of two feature sets \(\mathbf{X}\in\mathbb{R}^{N\times C}\) and \(\mathbf{Y}\in\mathbb{R}^{N\times C}\),
and \(\mathbf{H} = \mathbf{I} - \frac{1}{N}\mathbf{1}\mathbf{1}^{\top}\) is the centering matrix.
This formulation is invariant to isotropic scaling, allowing CKA to focus on the relative similarity structure encoded by the features.

We employ CKA to assess the registration of token representations rather than element-wise metrics such as MSE, as CKA compares the relational structure across samples, mitigating the influence of scale or magnitude differences and focusing on structural correspondence between features, which precisely aligns with the goal of registration.

\noindent \textbf{Detail on SSIM Loss.}
SSIM \cite{SSIM} was originally developed to compare the structural information between two images, \ie, the dependencies among spatial positions, which makes it particularly suitable for measuring spatial similarity between maps and thus for supervising registration. It also takes luminance and contrast into account. We compute the SSIM of all feature channels and average them as the final loss term:
\begin{equation}
\begin{aligned}
\mathcal{L}_{ssim}&(\mathbf{F}_1, \mathbf{F}_2)
= 1 - \mathrm{SSIM}(\mathbf{F}_1, \mathbf{F_2}) \\
&= 1 - \frac{1}{C}
\sum_{c=1}^{C}
\frac{
(2\mu_{\mathbf{F}_1^c}\mu_{\mathbf{F}_2^c} + C_1)\,(2\sigma_{\mathbf{F}_1^c \mathbf{F}_2^c} + C_2)
}{
(\mu_{\mathbf{F}_1^c}^2 + \mu_{\mathbf{F}_2^c}^2 + C_1)\,(\sigma_{\mathbf{F}_1^c}^2 + \sigma_{\mathbf{F}_2^c}^2 + C_2)
},
\end{aligned}
\end{equation}
where $\mu_{\mathbf{F}_i^c}$ and $\sigma_{\mathbf{F}_i^c}^2$ denote the \textit{local mean} and \textit{variance}
of channel $c$ in feature map $\mathbf{F}_i$,
$\sigma_{\mathbf{F}_1^c \mathbf{F}_2^c}$ denotes the \textit{local covariance} between the two feature maps,
and $C_1, C_2$ are small constants for numerical stability.
When combined, $\mathcal{L}_{ssim}$ focuses on spatial consistency across channels, whereas $\mathcal{L}_{cos}$ emphasizes semantic alignment between token vectors, making them mutually complementary for registration supervision.

\noindent \textbf{Dice Similarity Coefficient (DSC, Seg).}
For segmentation, DSC is defined as
\begin{equation}
\mathrm{DSC} = \frac{2|P \cap G|}{|P| + |G|},
\end{equation}
where \(P\) and \(G\) denote the predicted and ground-truth foreground sets.
\emph{Higher is better.}

\noindent \textbf{95th Percentile Hausdorff Distance (HD95, Seg).}
Hausdorff distance measures the boundary mismatch between prediction and ground truth:
\begin{equation}
\mathrm{HD}(P,G) = \max\left\{\sup_{p\in P}\inf_{g\in G} d(p,g), \sup_{g\in G}\inf_{p\in P} d(g,p)\right\},
\end{equation}
where \(d(\cdot,\cdot)\) is the Euclidean distance. We report the 95th percentile, denoted as HD95, to reduce sensitivity to outliers.
\emph{Lower is better.}

\noindent \textbf{Percentage of Correct Keypoints (PCK@0.1, LD).}
For landmark detection, a prediction \(\hat{p}_i\) is considered correct if its distance to the ground truth \(p_i\) is less than a threshold \(\alpha\) times a reference length \(L\) (image size, \ie, $L=256$). Let $N$ denote the number of annotated landmarks in the sample,
\begin{equation}
\mathrm{PCK@}\alpha = \frac{1}{N}\sum_{i=1}^N \mathbb{I}\!\left[\frac{\|\hat{p}_i - p_i\|_2}{L} < \alpha\right].
\end{equation}
We use \(\alpha=0.1\).
\emph{Higher is better.}

\noindent \textbf{Normalized Mean Squared Error (NMSE, LD).}
The average NMSE for landmark detection is
\begin{equation}
\mathrm{NMSE} = \frac{1}{N}\sum_{i=1}^N \frac{|\hat{p}_i - p_i|_2^2}{L^2},
\end{equation}
where $L = 256$ denotes the image size used for normalization.
\emph{Lower is better.}

\noindent \textbf{Mean Intersection over Union (mIoU, OD).}
The mIoU measures the overlap between predicted and ground-truth bounding boxes, and is defined as
\begin{equation}
\mathrm{mIoU} = \frac{1}{N} \sum_{i=1}^{N} \frac{|\hat{B}_i \cap B_i|}{|\hat{B}_i \cup B_i|},
\end{equation}
where $\hat{B}_i$ and $B_i$ denote the predicted and ground-truth bounding boxes, respectively, and $N$ is the number of matched objects.
\emph{Higher is better.}

\noindent \textbf{Normalized Mean Squared Error (NMSE, DE).}
The NMSE measures the relative pixel-wise regression accuracy for depth estimation, is defined as
\begin{equation}
\mathrm{NMSE} =
\frac{1}{N}\sum_{i=1}^{N}\frac{(\hat{d}_i - d_i)^2}{d_i^2}
,
\end{equation}
where $\hat{d}_i$ and $d_i$ denote the predicted and ground-truth depth values, respectively, and $N$ is the number of valid pixels in the depth map.
\emph{Lower is better.}

\noindent \textbf{Root Mean Squared Error (RMSE, OC).}
The RMSE measures the deviation between the predicted and ground-truth object counts, and is defined as
\begin{equation}
\mathrm{RMSE} = \sqrt{\frac{1}{N}\sum_{i=1}^{N} (\hat{c}_i - c_i)^2},
\end{equation}
where $\hat{c}_i$ and $c_i$ denote the predicted and ground-truth counts for the $i$-th sample. Here, $N$ denotes the number of samples in the evaluation set.
\emph{Lower is better.}

\noindent \textbf{Mean Absolute Error (MAE, OC).}
The MAE is defined as
\begin{equation}
\mathrm{MAE} = \frac{1}{N}\sum_{i=1}^{N} |\hat{c}_i - c_i|,
\end{equation}
where $\hat{c}_i$ and $c_i$ denote the predicted and ground-truth counts, respectively. $N$ denotes the number of samples in the evaluation set.
\emph{Lower is better.}

{
    \small
\begin{appendixbibliography}

\end{appendixbibliography}
}

\stopcontents[appendix]

\end{document}